\documentclass[twocolumn,10pt]{asme2ej}

\usepackage{graphicx} 
\usepackage{hyperref}   
\usepackage[]{enumitem}
\usepackage{multirow}
\usepackage[switch]{lineno} 
\usepackage{tikz}
\usepackage{fancyhdr}
\def\checkmark{\tikz\fill[scale=0.4](0,.35) -- (.25,0) -- (1,.7) -- (.25,.15) -- cycle;} 
\usepackage[english]{babel}
\nolinenumbers
\hypersetup{
	colorlinks=true,
	linkcolor=blue,
	citecolor=blue,
	urlcolor=blue,
}
\usepackage[square,numbers]{natbib}
\title{Multi-modal Machine Learning in Engineering Design: A Review and Future Directions}
\author{ Binyang ~Song\thanks{Address all correspondence for other issues to this author.}  \\
	\affiliation{Department of Mechanical Engineering\\
	Massachusetts Institute of Technology\\
	Cambridge, MA 02139 \\
        Email: binyangs@mit.edu}

}
\author{
    Rui Zhou \\
    \affiliation{
    Department of Mechanical Engineering\\
	Massachusetts Institute of Technology\\
	Cambridge, MA 02139\\
        Email: zhourui@mit.edu
    }
	 \\
}

\author{
    Faez ~Ahmed \\
    \affiliation{
	Department of Mechanical Engineering\\
	Massachusetts Institute of Technology\\
	Cambridge, MA 02139 \\
        Email: faez@mit.edu}
}

\begin{document}

\maketitle    

\begin{abstract}

{\it In the rapidly advancing field of multi-modal machine learning (MMML), the convergence of multiple data modalities has the potential to reshape various applications. This paper presents a comprehensive overview of the current state, advancements, and challenges of MMML within the sphere of engineering design. The review begins with a deep dive into five fundamental concepts of MMML:multi-modal information representation, fusion, alignment, translation, and co-learning. Following this, we explore the cutting-edge applications of MMML, placing a particular emphasis on tasks pertinent to engineering design, such as cross-modal synthesis, multi-modal prediction, and cross-modal information retrieval. Through this comprehensive overview, we highlight the inherent challenges in adopting MMML in engineering design, and proffer potential directions for future research. To spur on the continued evolution of MMML in engineering design, we advocate for concentrated efforts to construct extensive multi-modal design datasets, develop effective data-driven MMML techniques tailored to design applications, and enhance the scalability and interpretability of MMML models. MMML models, as the next generation of intelligent design tools, hold a promising future to impact how products are designed.}
\end{abstract}

\section{Introduction}
Our perception and interaction with the world involve multiple senses like sight, hearing, touch, and smell. These senses receive information through different \emph{mediums}, such as images and sound. A \emph{modality} refers to the \emph{medium} through which an object exists or is experienced. Common data modalities include~\cite{Bengio2012RepresentationPerspectives}:

\begin{enumerate}[label=(\arabic*)]
  \item Natural language (both spoken and written);
  \item Visuals (e.g., images, animations, and 3D shapes);
  \item Audios;
  \item Haptics;
  \item Smell and taste;
  \item Physiological signals (e.g., electrocardiogram, skin conductance, electroencephalograph);
  \item Other modalities (e.g., sensor data).
\end{enumerate}
In machine learning (ML), a modality refers to a certain data format in which information is stored and input to ML models. An ML model is multi-modal when it works with multiple modalities. Since we often rely on multiple media to perform real-world tasks, multi-modal machine learning (MMML) is essential to replicate human-like capabilities for these tasks.

\subsection{Definition of Multi-Modal Machine Learning}
Traditional ML models typically rely on a single data modality, which we refer to as unimodal ML. Unimodal ML focuses on learning latent representations specific to each modality for various downstream tasks. In contrast, MMML involves training models on multiple forms of data. MMML presents challenges as it requires understanding and integration of various data forms. However, MMML also offers power by leveraging the complementarity, alignment, and redundancy of multi-modal data, enabling a more comprehensive understanding of instances and facilitating cross-modal problem-solving. In this paper, we define an \emph{instance} as a data point represented by multiple modalities in a multi-modal dataset. A \emph{feature} represents a vectorized latent representation of input data at different learning stages, capturing real-world data characteristics. Within this framework, complementarity refers to the synergy between information components from different modalities in describing an instance. Alignment represents the correspondence between the information components across modalities. Redundancy indicates that information components from different modalities have the same meaning, enhancing the robustness of the models learning from them.

\textcolor{black} {Following the definition of optimization problems, we refer to ML involving two or three data modes as MMML and that involving more than three data modes as ``many-modal learning''~\cite{10.1115/1.4039450}. Currently, MMML primarily focuses on bimodal machine learning with two data modes. As the internet connects an increasing number of cyber-physical systems, MMML is expected to incorporate more than two data modes and further evolve into many-modal learning, enabling models to utilize diverse data from various sensors. Since algorithms designed for multi-objective problems struggle with the larger dimensionality of many-objective problems~\cite{10.1115/1.4039450}, incorporating more data modes will probably introduce new challenges as well.}

MMML has the potential to revolutionize ML research and applications across domains by enabling models to leverage multiple data modes. For instance, MMML can enhance diagnosis accuracy by integrating medical records, images, and patient-reported symptoms in healthcare. The impact of MMML extends beyond industries, driving innovation and efficiency in diverse sectors and ultimately boosting the US economy. However, MMML adoption in engineering design applications lags behind. This paper reviews MMML's existing work, explores methods and applications, and highlights key challenges in its engineering design implementation.

\subsection{Multi-modality in Design Representation}
Design researchers have made notable progress in utilizing state-of-the-art deep learning (DL) models for design synthesis~\cite{Zhu2023BiologicallyTransformers, Zhu2023GenerativeGeneration, Nobari2021PcDGAN:Design, LUO2021106873}, evaluation~\cite{Song2023ATTENTION-ENHANCEDEVALUATIONS, 10.1115/1.4049895}, and optimization~\cite{Nobari2021Range-GAN:Synthesis}. The effectiveness of these models heavily relies on the representation of the learned designs~\cite{Song2022AssessingPrediction}. To support formal and functional reasoning, a representation framework with ample capacity is essential~\cite{Gero1990DesignDesign}. Distinct representation modes, such as rough sketches, realistic renderings, and detailed 3D models~\cite{Tseng2011HowTask, Haggman2015ConnectionsDesign, Tsai2017HowCAD}, require varying design resources in terms of skill and time, resulting in varying levels of resolution and fidelity. The choice of design representation mode significantly influences human perception, creative design performance, and the occurrence of design fixation~\cite{Haggman2015ConnectionsDesign, Tsai2017HowCAD, Purcell1998DrawingsPsychology, Ullman1990TheProcess, Chang2016EffectsAbilities, Atilola2016TheTrees}. 

Hand-drawn sketches serve as a universal language to communicate initial ideas in early design stages, enabling efficient exploration at different abstraction levels while preserving ambiguity for innovation and creativity~\cite{Purcell1998DrawingsPsychology, Hannibal2016AnDesign, Atilola2015RepresentingSketches}. The digital revolution and advancements in computer-aided design (CAD) software offer expanded options for design representation. Two-dimensional (2D) and three-dimensional (3D) representations are commonly employed for generating interactive and complex solutions during the detailed design phase~\cite{Hannibal2016AnDesign}. The adoption of CAD facilitates deep exploration rather than broad exploration~\cite{Ullman1990TheProcess}. Renderings and view silhouettes of 3D CAD representations offer a simplified way to convey designs~\cite{Reid2013ImpactJudgment}. Physical prototypes offer a distinct approach to design representation, enabling clear and accurate interpretation of 3D spatial relationships~\cite{Hannibal2016AnDesign, Yang2005AOutcome}. Clay modeling, commonly used in automobile styling design, brings designs to life and facilitates intuitive design, refinement, and improvement. In the absence of physical prototypes, designers rely on photographs or textual descriptions to communicate designs instead~\cite{Atilola2015RepresentingSketches, McKoy2020InfluenceGeneration}. Alternatively, design concepts can be represented by categorical or continuous parametric data~\cite{Grace2014Data-intensiveSurprise, Nomaguchi2019AssessingDesign}. To enhance information exchange during conceptual design, researchers have explored theoretical representation methods, such as function model~\cite{Wood2001ProductDevelopment}, functional decomposition and morphology~\cite{Ulrich2000ProductDevelopment}, problem solution network~\cite{Fiorineschi2018IssuesAssessments}, and morphological charts~\cite{Fiorineschi2018IssuesAssessments}.

In engineering design, multiple data modes are combined to represent designs, such as sketches with textual descriptions or prototypes with parametric and verbal information. The choice of design representation evolves across design stages, transitioning from abstract formats to detailed formats. Employing different representation modes at different stages enables a smooth exploration process~\cite{Veisz2012Computer-aidedStudy, BABAPOUR2014MediaEducation}. While multiple modalities are commonly used in design practice, most ML applications supporting design research focus on unimodal approaches. MMML holds great potential for comprehensive AI-driven design comprehension. This could lead to more accurate design evaluation, reducing the manual effort required for assessment. Additionally, MMML enables cross-modal syntheses, such as text-to-image and image-to-shape generation, facilitating design exploration.

\subsection{The Scope of This Paper}
\begin{figure*}[!ht]
    \centering
    \includegraphics[width=16cm]{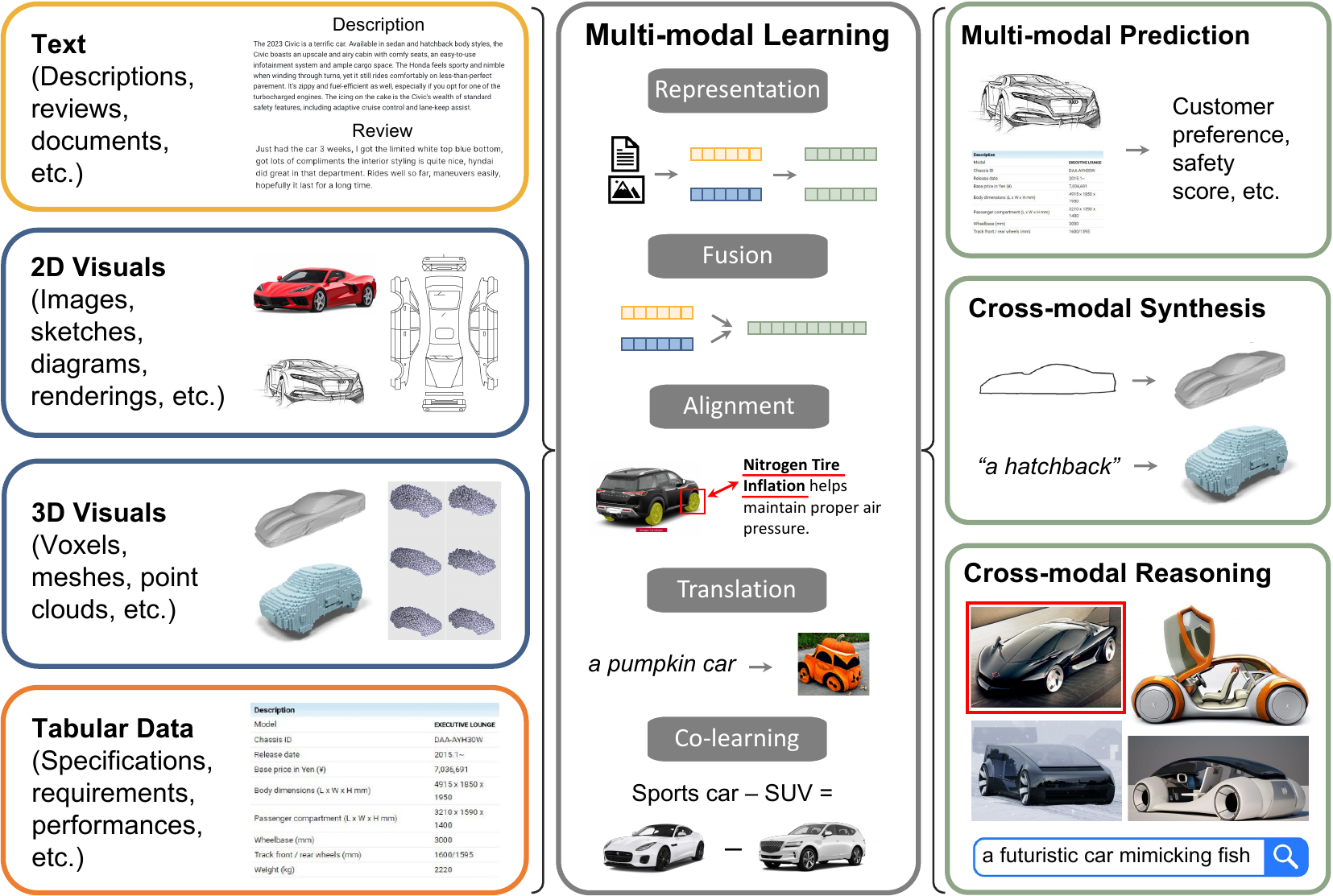}
    \caption{An illustration of the scope of this paper, using car designs as a prime example.}
    \label{fig:overview}
\end{figure*}

This paper presents a technical review of the fundamental concepts and applications of MMML. \textcolor{black} {As the scope of this paper depicted in Figure~\ref{fig:overview}, text, 2D and 3D visuals, and parametric data, commonly used as design representations, are exploited for MMML in engineering design. The review delves into five foundational MMML concepts: multi-modal information representation, fusion, alignment, translation, and co-learning. These techniques form the backbone of a myriad of applications such as multi-modal prediction, cross-modal synthesis, and cross-modal reasoning.} Future directions and challenges of MMML adoption in engineering design are also discussed. \textcolor{black} {Unlike prior MMML review papers in other domains~\cite{Baltrusaitis2019multimodalTaxonomy, Zhang2019multimodalApplications, Cui2022DeepReview}, this paper focuses on MMML techniques and applications potentially supporting primary design tasks like design synthesis and evaluation and design knowledge extraction, while giving little attention to less relevant topics like audio-visual speech recognition and speech synthesis. Furthermore, this paper distinguishes itself from a review paper published in the Journal of Mechanical Design~\cite{zhenghui2022} by reviewing and discussing the fundamental concepts and techniques underlying MMML and covering its applications more comprehensively. Compared to all prior review papers, this paper covers more newly emerging techniques like denoising diffusion models (DDMs) and their corresponding conditioning mechanisms~\cite{Dhariwal2021DiffusionSynthesis, Nichol2021GLIDE:Models, Kim2021DiffusionCLIP:Manipulation}.} It is organized as follows: Section 2 reviews the fundamental concepts of MMML. Section 3 examines MMML applications relevant to engineering design. Section 4 addresses the challenges and opportunities in applying MMML to engineering design. Finally, Section 5 provides a summary to conclude the paper.

\section{Fundamental Concepts of Multi-modal Machine Learning}
This section reviews the fundamental concepts of MMML, including representation, fusion, alignment, translation, and co-learning. These elements enable MMML to handle data heterogeneity and exploit data complementarity, alignment, and redundancy for various cross-modal applications. These are the key ideas differentiating MMML from unimodal ML and supporting various multi-modal applications. In MMML, multi-modal data can be categorized as parallel, non-parallel, or hybrid~\cite{Baltrusaitis2019multimodalTaxonomy}. Parallel data comprises associated instances from multiple modalities, such as images with captions. Non-parallel data does not require direct associations between modalities but shares common categories, such as images from different categories and corresponding Wikipedia pages ~\cite{Frome2013DeViSE:Model}. Hybrid data involves modalities linked through a pivot modality, where each modality is partially associated with the pivot modality, such as different languages connected via English as the pivot language~\cite{Rajendran2015BridgeLearning}. We now turn to the discussion of multi-modal representation.

\subsection{Representation}
Representation is the basis of information processing for both humans and computers. In the context of MMML, representing multi-modal data involves learning vector representations that capture data complementarity, alignment, and redundancy across modalities. Typically, unimodal representations are first learned separately from each modality and then fused to obtain multi-modal representation.

\subsubsection{Different Representation Forms} \label{rep}
\begin{figure}[!ht]
    \centering
    \includegraphics[width=8cm]{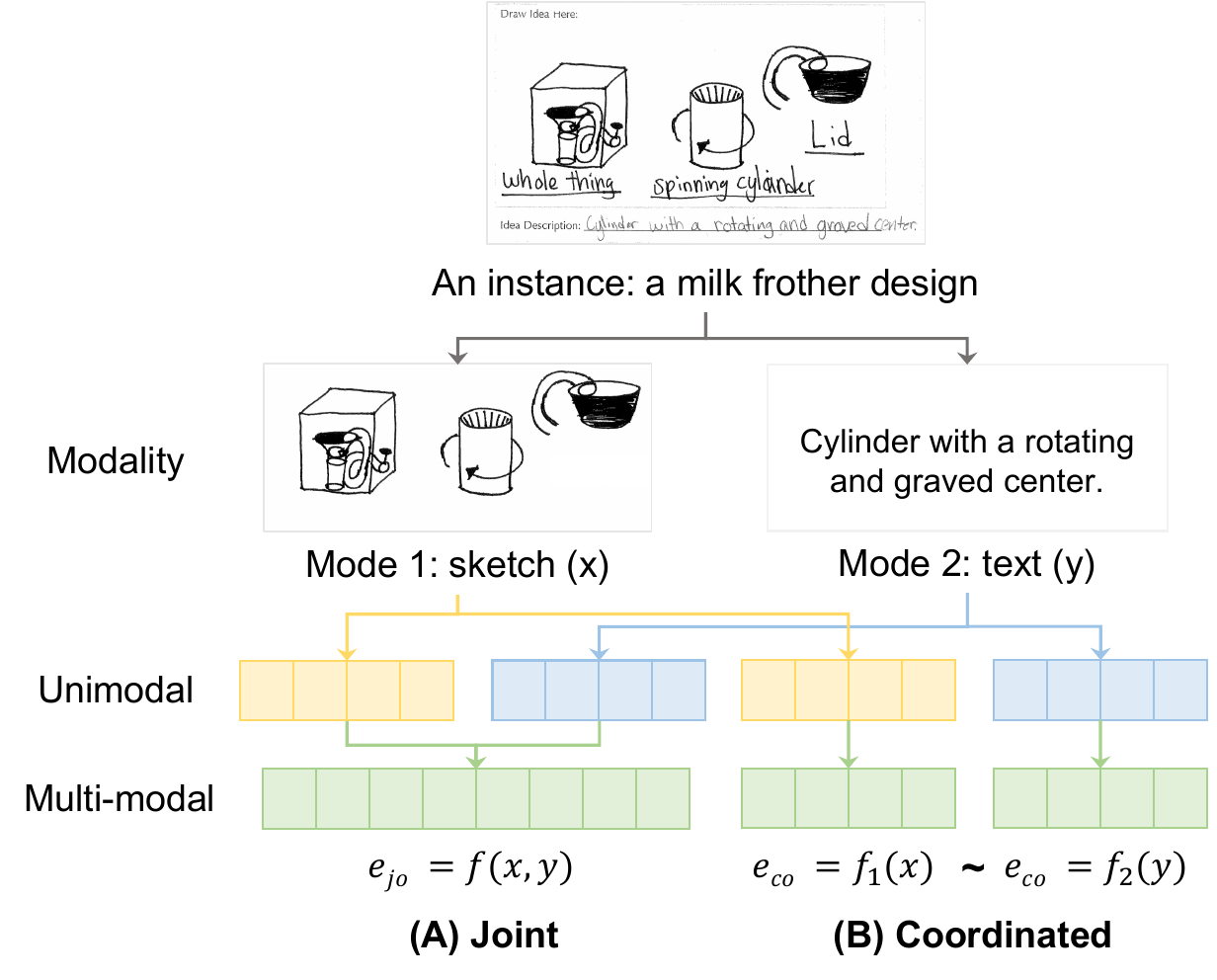}
    \caption{Joint and coordinated representations demonstrated by a sketch-text instance. While joint representations are projected to the same space using all modalities as input, coordinated representations exist in their own space but are coordinated.}
    \label{fig:repre}
\end{figure}

Multi-modal representations can be classified into two categories: joint representation and coordinated representation~\cite{Baltrusaitis2019multimodalTaxonomy}. Joint representation involves fusing multiple unimodal representations into a single multi-modal representation, enabling modalities to complement each other (Figure~\ref{fig:repre}-A). This approach is commonly used for multi-modal prediction tasks~\cite{Srivastava2012multimodalMachines, DucTuan2021multimodalDetection, Song2022HEYAnd, Yuan2022LeveragingModel}, and various fusion methods will be discussed in the following subsection. When high-quality labels are available, joint representations can be learned through supervised learning with rich intra-modal and cross-modal interactions~\cite{Song2022HEYAnd, Nguyen2018Multi-taskRepresentation}. Otherwise, we can employ self-supervised approaches to learn joint representations via pre-training tasks (e.g., masked content prediction)~\cite{Li2020Unicoder-VL:Pre-Training, Su2019VL-BERT:Representations, Li2019VisualBERT:Language, Alberti2019FusionAnswering, Sun2019VideoBERT:Learning} or using generative models~\cite{Srivastava2012multimodalMachines, Ngiam2011multimodalLearning, Silberer2014LearningAutoencoders, Feng2014Cross-modalAutoencoder}. On the other hand, coordinated representations involve learning multiple coordinated unimodal representations learned from associated modalities (Figure~\ref{fig:repre}-B). This is achieved by projecting different modalities to a common subspace and maximizing similarity~\cite{Radford2021LearningSupervision}, correlation~\cite{Andrew2013DeepAnalysis, Yang2017DeepData, Feng2014Cross-modalAutoencoder}, mutual information~\cite{Bachman2019LearningViews}, or agreement~\cite{Zhang2020ContrastiveText} between the associated unimodal representations. This approach captures correlations and mutual information across modalities supporting tasks like cross-modal IR~\cite{Kiros2014UnifyingModels, Huang2013LearningData, Feng2014Cross-modalAutoencoder} and synthesis~\cite{Kiros2014UnifyingModels}. Coordinated representations can be learned at the instance level or finer levels (e.g., image or sentence fragments) to facilitate fine-grained reasoning~\cite{Karpathy2014DeepDescriptions, Karpathy2014DeepMapping, Wu2019UnifiedRepresentations, Plummer2015Flickr30kModels}. Coordinated representations are typically learned in a task-agnostic and self-supervised way, where different modalities can serve as the ``supervision'' of each other~\cite{Radford2021LearningSupervision, Andrew2013DeepAnalysis, Yang2017DeepData, Bachman2019LearningViews, Zhang2020ContrastiveText}. While joint representations apply to two or more modalities, coordinated representations mostly focus on the relationship between two modalities.

Several pre-trained, generalizable multi-modal representations have been released to facilitate effective representation learning for various tasks. The majority of these pre-trained models focus on the textual and visual modalities due to the availability of paired textual and visual data. The joint representations are commonly learned using transformer-based models, such as Unicoder-AL~\cite{Li2020Unicoder-VL:Pre-Training}, VL-BERT~\cite{Su2019VL-BERT:Representations}, VisualBERT~\cite{Li2019VisualBERT:Language}, and B2T2~\cite{Alberti2019FusionAnswering}, LXMERT~\cite{Tan2019LXMERT:Transformers}, ViLBERT~\cite{Lu2019ViLBERT:Tasks}, and OmniNet~\cite{Pramanik2019OmniNet:Learning}. These models leverage attention mechanisms to capture cross-modal alignment and are frequently used for tasks like VQA~\cite{Li2020Unicoder-VL:Pre-Training} and image captioning~\cite{Su2019VL-BERT:Representations}. On the other hand, contrastive language-image pre-training (CLIP)~\cite{Radford2021LearningSupervision} is a popular coordinated representation. CLIP is pre-trained to predict which captions go with which images, maximizing the similarity between the associated image and caption representations. CLIP representations are transferable to various tasks and competitive even without dataset-specific training, making it a widely used representation for cross-modal synthesis tasks. Another coordinated representation, recently proposed, is contrastive image-shape pre-training (CISP)~\cite{Sbrolli2022IC3D:Generation}. CISP converts images and shapes to patch embeddings using 2D and 3D convolutions and encodes them separately with transformer-based encoders. It supports cross-modal syntheses between 2D images and 3D shapes.

The process of learning joint representations involves a step to fuse multiple unimodal representations. In the next section, we will review the diverse fusion methods, each exploiting unimodal features from varying perspectives.

\subsection{Multi-modal Fusion}
\begin{figure*}[!ht]
    \centering
    \includegraphics[width=15cm]{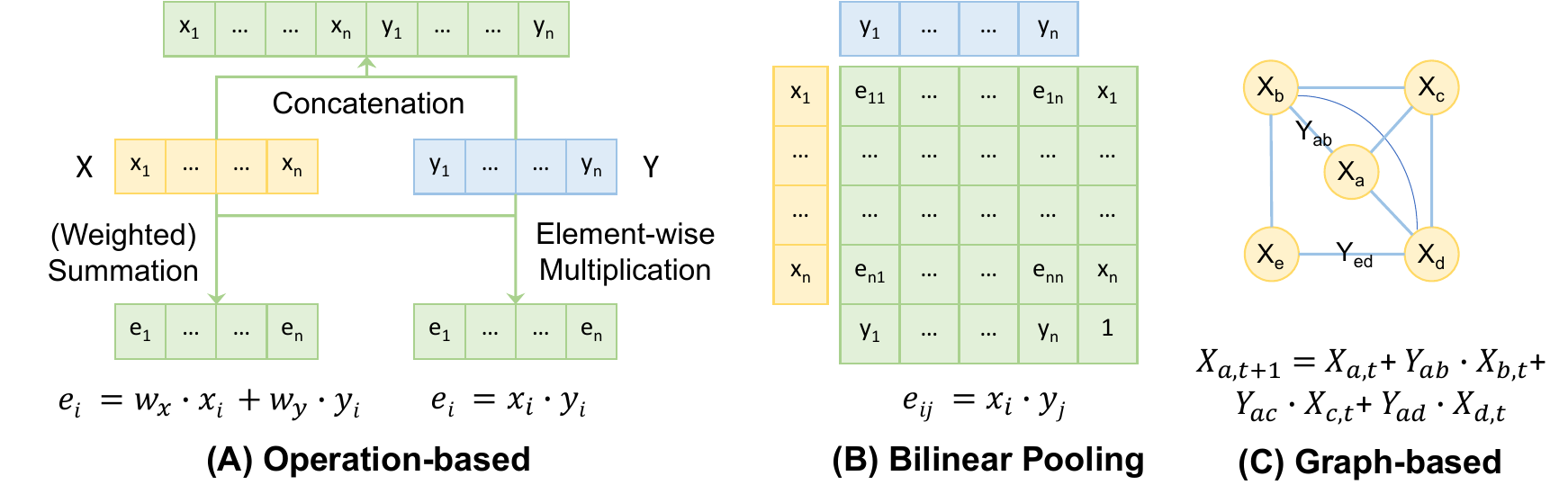}
    \caption{\textcolor{black} {Different multi-modal information fusion methods with X and Y standing for two unimodal representations: A) Operation-based fusion, where X and Y are fused through concatenation, weighted summation, and element-wise multiplication, respectively; B) Bilinear pooling fusion, where X and Y are fused by calculating their outer product; C) Graph-based fusion, where X and Y are fused through message passing between nodes.}}
    \label{fig:fusion}
\end{figure*}

Multi-modal fusion involves the combination of information from multiple modalities for prediction tasks like classification or regression. It enhances prediction quality through three key aspects: 1) Improved robustness due to multi-modal information redundancy; 2) Increased accuracy by leveraging multi-modal information complementarity; 3) The ability to make predictions in the absence of a certain modality~\cite{Baltrusaitis2019multimodalTaxonomy}. Given the dominance of DNNs in various tasks in recent years, our focus is on multi-modal information fusion within DL. Based on previous definitions~\cite{Baltrusaitis2019multimodalTaxonomy, Zhang2019multimodalApplications, Cui2022DeepReview}, we classify multi-modal information fusion into three types: operation-based, bilinear pooling, and graph-based fusion, illustrated in Figure~\ref{fig:fusion}.

\subsubsection{Operation-based Fusion}
The operation-based approaches integrate unimodal representations using simple operations, such as concatenation~\cite{Nojavanasghari2016DeepPrediction, Anastasopoulos2019NeuralFeatures, Vielzeuf2019CentralNet:Fusion, 10.1115/1.4054001} (Figure~\ref{fig:fusion}-A), or element-wise averaging~\cite{Shutova2016BlackFeatures}, multiplication~\cite{Cao2016DeepRetrieval}, linear combination~\cite{Anastasopoulos2019NeuralFeatures, Vielzeuf2019CentralNet:Fusion, Sikka2013MultipleWild}, and majority voting~\cite{Morvant2014MajorityFusion}. The pre-learned unimodal representations need to be dimensionally consistent and appropriately ordered for element-wise operations~\cite{Perez-Rua2019MFAS:Search}. The fusion can be implemented early, late, or in a hybrid manner during the learning stages. Early fusion integrates low-level unimodal features, capitalizing on inter-modality correlations~\cite{Perez-Rua2019MFAS:Search}. Despite its simplicity, it can lead to overfitting due to high dimensionality, particularly with limited training data. Late fusion, on the other hand, merges high-level features~\cite{Morvant2014MajorityFusion}, facilitating intra-modality interactions. It allows for higher flexibility of the unimodal models and copes well with missing modalities. Hybrid fusion, combining the benefits of both early and late fusion, adds complexity due to its intricate mechanisms~\cite{Anastasopoulos2019NeuralFeatures, Zhou2019EffectiveDiagnosis}. The optimal fusion architecture is typically determined through experimentation. Researchers have explored using reinforcement learning~\cite{Zoph2016NeuralLearning} or surrogate models~\cite{Perez-Rua2019MFAS:Search} to identify promising fusion architectures.

\subsubsection{Bilinear Pooling Fusion}
Bilinear pooling fusion (Figure~\ref{fig:fusion}-B), also known as tensor-based~\cite{Cui2022DeepReview} or bilinear model-based fusion~\cite{Tenenbaum2000SeparatingModels}, integrates unimodal feature vectors by calculating their outer~\cite{Tenenbaum2000SeparatingModels} or Kronecker~\cite{Zadeh2017TensorAnalysis, Chen2019PathomicPrognosis} product. This approach captures high-order multiplicative interactions among modalities, yielding highly expressive multi-modal representations~\cite{Tenenbaum2000SeparatingModels, Zadeh2017TensorAnalysis}. Feature vectors are often extended with an additional value of one to preserve unimodal features~\cite{Zadeh2017TensorAnalysis}. Despite its advantages, handling large tensors resulting from high-dimensional inputs can be computationally demanding or impractical. This prompts the adoption of low-dimensional approximations, i.e., decomposing high-dimensional tensors into multiple low-dimensional tensors, for balancing expressivity and computational efficiency~\cite{Kim2022HadamardPooling, Yu2017multi-modalAnswering, Yu2017BeyondAnswering, Gao2015CompactPooling, Fukui2016multimodalGrounding, Ben-Younes2017MUTAN:Answering, Tucker1966SomeAnalysis, Ben-Younes2019BLOCK:Detection}.

\subsubsection{Graph-based Fusion}
The graph-based methods utilize graphs to model relations between elements to fuse multi-modal features (Figure~\ref{fig:fusion}-C). In such graphs, nodes represent instances, while edges signify inter-instance relationships, with node and edge embeddings respectively carrying modality-specific information~\cite{Jiang2022DeepClassification}. Graph neural networks (GNNs) are utilized to learn these graphs, which update the embeddings through information passing to fuse information from different modalities. Studies show this approach, used for fusing image and non-image features, surpasses simple concatenation-based fusion~\cite{Parisot2018DiseaseDisease, Cao2021UsingData}.

\textcolor{black} {In addition to the aforementioned methods, attention mechanisms, powerful tools for aligning features across modalities, are also common in multi-modal information fusion~\cite{Song2023ATTENTION-ENHANCEDEVALUATIONS, DucTuan2021multimodalDetection, Yuan2022LeveragingModel} and will be separately discussed in the next subsection about multi-modal information alignment.}

\subsection{Alignment}
Alignment is the process of correlating information elements from multiple modalities. In DL, it does not necessitate explicit feature alignment or supervision~\cite{Baltrusaitis2017multimodalTaxonomy}. In recent years, attention mechanisms have become a favored method for aligning multi-modal features~\cite{Zhang2019multimodalApplications,Cui2022DeepReview}. They excel in modeling dependencies between a focal feature (i.e., query) and all candidate features (i.e., key), which allows more attention to be paid to more relevant features (i.e., value)~\cite{Vaswani2017AttentionNeed, Graves2014NeuralMachines}. Multi-head attention, allowing multiple focuses to draw attention simultaneously, provides more comprehensive retention of important information~\cite{Vaswani2017AttentionNeed}. In MMML, cross-attention uses focal features from one modality to locate relevant features from another, enhancing model performance and interpretability~\cite{Zhang2019multimodalApplications, Bahdanau2014NeuralTranslate}. Figure~\ref{fig:alignment} illustrates different attention mechanisms aligning textual descriptions and sketches.

\subsubsection{Customized Attention}
\begin{figure*}[!ht]
    \centering
    \includegraphics[width=15cm]{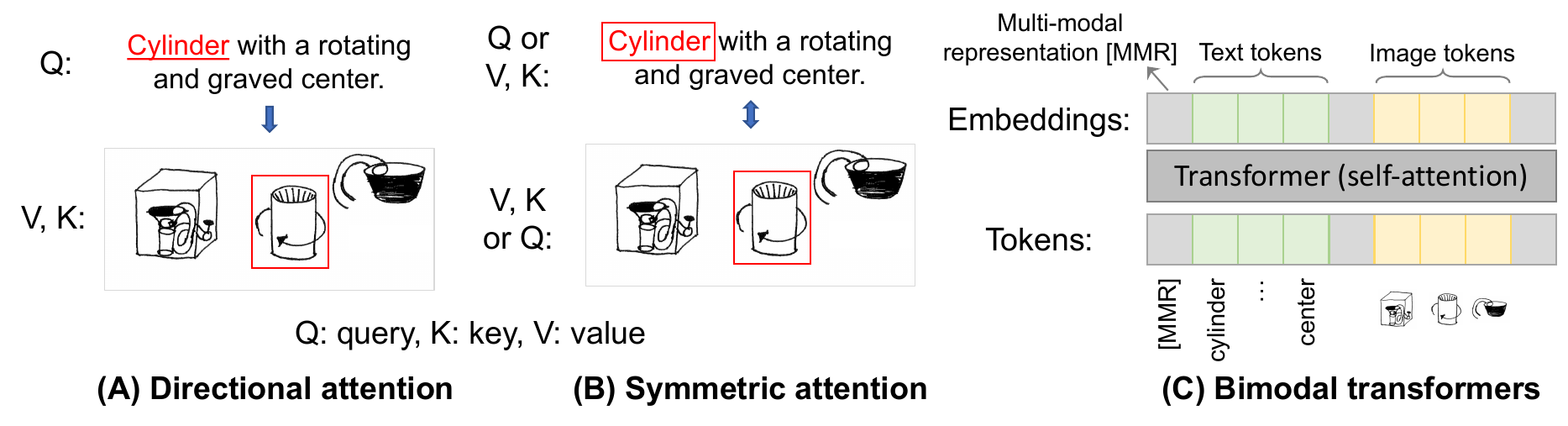}
    \caption{\textcolor{black} {Different alignment mechanisms demonstrated by a sketch-text instance: A) Directional attention, which uses text queries to attend to different image regions; A) Symmetric attention, which uses text and image queries to attend to different image regions and text tokens, respectively; C) Bidirectional transformers, where the image and text tokens are weighted through self-attention embedded in transformers.}}
    \label{fig:alignment}
\end{figure*}

Attention mechanisms can be directional or symmetric for different DL tasks, as shown in Figure~\ref{fig:alignment}. Examples of directional attention (Figure~\ref{fig:alignment}-A) include visual attention, which attends to semantic features using visual features for VQA~\cite{Zhu2016Visual7W:Images, Shih2015WhereAnswering} and image captioning~\cite{Xu2015AskAnswering, Anderson2017Bottom-UpAnswering}, and semantic attention, which attends to visual features using semantic features for text-to-image synthesis~\cite{Mansimov2015GeneratingAttention, Xu2018AttnGAN:Networks, Li2019Object-drivenTraining}. To improve cross-modal interaction reasoning, attention mechanisms can be made bi-directional and symmetric between two modalities (Figure~\ref{fig:alignment}-B)~\cite{Nam2017DualMatching, Lu2016HierarchicalAnswering, Osman2018DualAnswering}. This approach has demonstrated its effectiveness in multi-modal classification~\cite{DucTuan2021multimodalDetection}, regression~\cite{Song2022HEYAnd}, IR~\cite{Lu2016HierarchicalAnswering, Nam2017DualMatching}, and VQA~\cite{Schwartz2017High-OrderAnswering}. Stacking multiple attention layers or stages can capture richer alignment information and promote progressive reasoning, as seen in applications to VQA~\cite{Yang2015StackedAnswering, Fan2018StackedReasoning, Xiong2016DynamicAnswering, Xu2015AskAnswering, Ren2015FasterNetworks, Anderson2017Bottom-UpAnswering, Lu2017Co-attendingAnswering} and text-to-image synthesis~\cite{Rombach2021High-ResolutionModels}.

\subsubsection{Multi-modal Transformer}
Transformers~\cite{Vaswani2017AttentionNeed}, built with attention and feedforward layers, excel in learning sequential data, notably long sequences. Their successful implementation in unimodal natural language processing (NLP) has prompted researchers to extend their applications to multi-modal domains. Models like Unicoder-AL~\cite{Li2020Unicoder-VL:Pre-Training}, VL-BERT~\cite{Su2019VL-BERT:Representations}, VisualBERT~\cite{Li2019VisualBERT:Language}, and B2T2~\cite{Alberti2019FusionAnswering} leverages transformers to learn semantic, visual, and contextualized multi-modal embeddings from input visual and textual tokens (Figure~\ref{fig:alignment}-C), using pre-training tasks like masked token prediction. Others, like LXMERT~\cite{Tan2019LXMERT:Transformers} and ViLBERT~\cite{Lu2019ViLBERT:Tasks}, take a two-stream approach, extracting unimodal features from tokens with transformers, then using cross-modal attention to align these features. Data2Vec~\cite{data2vec} further extends transformers to multi-modal contexts, aiming for a universal multi-modal self-supervised learning scheme to produce contextualized latent representations of full input data. 

In addition to attention mechanisms, alternative methods for weighing data elements include multi-modal residual networks~\cite{Kim2016multimodalQA}, gated multi-modal units~\cite{Arevalo2017GatedFusion}, and dynamic parameter layers~\cite{Noh2015ImagePrediction}. For more details, interested readers can refer to the respective papers. While coordinated representations (Sec~\ref{rep}) aim at capturing global correlations at the instance level, attention mechanisms enable the learning of local and finer-grained alignment at the feature level. Capturing cross-modal alignment at different levels powers cross-modal translation, which we will discuss next.

\subsection{Translation}
\begin{figure*}[!ht]
    \centering
    \includegraphics[width=16cm]{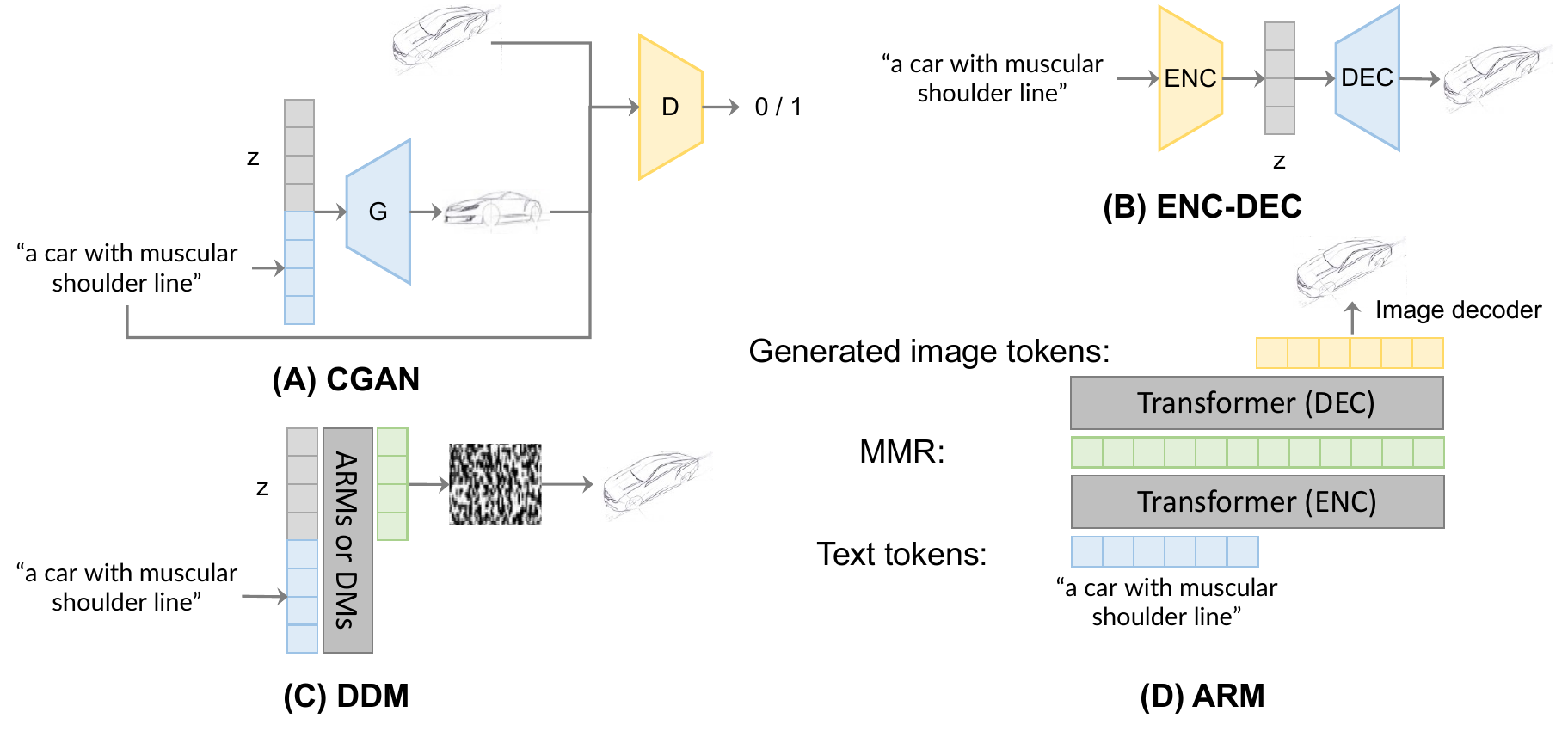}
    \caption{\textcolor{black} {Different text-to-sketch translation models: A) Conditional generative adversarial network (CGAN), where text embeddings are concatenated with sampled noise as conditioning; B) Multi-modal variation auto-encoder (VAE), where text embeddings are decoded to sketches; C) Denoising diffusion model (DDM), where text embeddings are fused with sampled noise as conditioning; D) Auto-regressive model (ARM), where a transformer encoder learns text embeddings, which are decoded to sketches using a transformer decoder.}}
    \label{fig:synthesis}
\end{figure*}

Cross-modal translation involves synthesizing information from one modality into another. This task requires understanding source mode information and the capability to generate corresponding signals in the target mode. In ML, cross-modal translation includes generating non-existent samples in one modality according to input from another modality and modifying existing samples based on input from another modality. Despite advancements in deep generative models (DGMs), this remains challenging. This subsection examines two classes of cross-modal translation models, generative adversarial networks (GANs) and likelihood-based models. Unlike GANs' adversarial approach, likelihood-based models learn data distributions using likelihood functions and include VAEs, DDMs, autoregressive models (ARMs), and flow-based models. We will discuss different cross-modal translation models using text-to-car sketch synthesis as an example (Figure~\ref{fig:synthesis}) and explore evaluation metrics for synthesis quality.

\subsubsection{Conditional Generative Adversarial Networks}
Conditional generative adversarial networks (CGANs) are popular for text-to-image translation (Figure~\ref{fig:synthesis}-A). GANs comprise a generator and a discriminator, trained competitively for high-quality image synthesis~\cite{Tolstikhin2014GenerativeNetworks}. In CGANs, both components are conditioned on text prompts~\cite{Mirza2014ConditionalNets, Reed2016GenerativeSynthesis}. The generator combines random noise with embeddings of text prompts as conditioning, while the discriminator evaluates the pairing of the generated images and the text prompts. CGANs rely on parallel data for training. Researchers have enhanced image quality and text-image alignment by conditioning augmentation~\cite{Zhang2016StackGAN:Networks}, stacking multiple CGANs~\cite{Zhang2016StackGAN:Networks, Zhang2019StackGAN++:Networks, Xu2018AttnGAN:Networks}, semantic attention~\cite{Xu2018AttnGAN:Networks}, dynamic memory~\cite{Zhu2019DM-GAN:Synthesis}, or adding new loss modules~\cite{Zhang2019StackGAN++:Networks, Xu2018AttnGAN:Networks, Zhu2019DM-GAN:Synthesis, Zhang2018PhotographicNetwork, Dash2017TAC-GANNetworkb, Cha2018AdversarialSynthesis, Qiao2019MirrorGAN:Redescription}. CGANs can also be conditioned on semantic concept layouts or scene graphs, enabling synthesis of complex images with multiple objects~\cite{Reed2016LearningDraw, Zhao2018ImageLayout, Hinz2019GeneratingLocations, Hong2018InferringSynthesis, Johnson2018ImageGraphs}.

\subsubsection{Multi-modal Variational Autoencoders}
Multi-modal VAEs, comprising encoder-decoder pairs (Figure~\ref{fig:synthesis}-B), first encode source instances into latent multi-modal representations, then generate target instances from these representations, bridging the source and target modes for image captioning~\cite{Mao2014Deepm-RNN} and text-to-image~\cite{Mansimov2015GeneratingAttention, vandenOordDeepMind2017NeuralLearning} or -geometry~\cite{Sanghi2021CLIP-Forge:Generation} translation. Multi-modal VAEs can be trained on parallel data~\cite{Kiros2014UnifyingModels} or nonparallel data with pre-trained coordinated representations like CLIP~\cite{Sanghi2021CLIP-Forge:Generation}. Multi-modal VAEs can enhance synthesis quality when integrated with GANs~\cite{Shetty2017SpeakingTraining}. The choice of encoder and decoder architectures should align with the source and target modes; convolutional neural networks (CNNs) for image encoding or generation~\cite{Ajit2020ANetworks, Li2021AProspects}, recurrent neural networks (RNNs) and distributed models~\cite{Fathi2018DeepProcessing, Mikolov2013DistributedCompositionality, Yagcioglu2015ACaptioning} for language encoding or generation, and transformer-based models for both~\cite{Cordonnier2019OnLayers, Dosovitskiy2020AnScale, Wang2022End-to-EndCaptioning, Kalyan2021AMMUSProcessing}. While multi-modal VAEs offer diverse sampling capabilities and simpler training procedures compared to CGANs, it comes with the trade-off of lower fidelity.

\subsubsection{Denoising Diffusion Models}
DDMs are latent variable DGMs featuring a forward diffusion process that adds noise to input data and a denoising process that removes noise to regenerate input data~\cite{Sohl-Dickstein2015DeepThermodynamics} (Figure~\ref{fig:synthesis}-C). These processes can be parameterized by different time-dependent Gaussian transitions, leading to various DDMs, such as denoising diffusion probabilistic models (DDPMs)~\cite{Ho2020DenoisingModels}, denoising diffusion implicit models (DDIMs)~\cite{Song2020DenoisingModels}, and score-based models~\cite{Song2020Score-BasedEquations, Vahdat2021Score-basedSpace}. While DDPMs and DDIMs maximize the variational lower bound of the likelihood functions representing the data distributions, score-based models minimize losses of matching time-dependent gradients, i.e., scores. DDMs excel in conditional and unconditional image~\cite{Song2020Score-BasedEquations, Ho2020DenoisingModels, Nichol2021GLIDE:Models, Kim2021DiffusionCLIP:Manipulation} and geometry~\cite{Luo2021DiffusionGeneration, Zhou20213DDiffusion, Zeng2022LION:Generation} synthesis. To boost data efficiency, latent diffusion models (LDMs) compress high-dimensional data samples into low-dimensional latent spaces~\cite{Rombach2021High-ResolutionModels, Liu2019Point-VoxelLearning}. DDMs can be conditioned for cross-modal image or shape syntheses through two primary methods: parameter-based and embedding-based approaches. The parameter-based approach modifies Gaussian transition parameters based on classifier guidance~\cite{Dhariwal2021DiffusionSynthesis}, similarity-based guidance~\cite{Nichol2021GLIDE:Models,Kim2021DiffusionCLIP:Manipulation}, or classifier-free guidance~\cite{Ho2022Classifier-FreeGuidance}, steering the reverse diffusion process. While classifier or similarity-based guidance uses separate, noise-aware modules to predict classes or similarity scores of synthesized samples, classifier-free guidance simply uses the embedding of input data, exhibiting superior effectiveness~\cite{Nichol2021GLIDE:Models, Kim2021DiffusionCLIP:Manipulation, Rombach2021High-ResolutionModels}. The conditioning effect can be enhanced with cross-model attention mechanisms\cite{Rombach2021High-ResolutionModels}. The embedding-based approach incorporates encoded conditioning information into sampled noise, optimizing sample diversity while maintaining realism and cross-modal correspondence~\cite{Nichol2022Point-E:Prompts, Ramesh2022HierarchicalLatents}. Recently, DDMs have gained traction for cross-modal translation~\cite{Dhariwal2021DiffusionSynthesis}.

\subsubsection{Autoregressive Models}
Deep generative ARMs use the chain rule of conditional probability to learn sequence distributions, predicting each sequence element from preceding ones. ARMs are mainly used for tokenized sequential data syntheses, such as text~\cite{Mao2015GenerationDescriptions, Vinyals2014ShowGenerator, Rohrbach2015TheDescription} and quantized images~\cite{Yu2022ScalingGeneration,Ding2021CogView:Transformers}. Common ARMs for cross-modal text synthesis include RNNs, LSTMs, and transformer-based models, with the latter also employed for image synthesis~\cite{Desai2020VirTex:Annotations, BulentSariyildiz2020LearningAnnotations}. ARMs can also take multi-modal tokens as input and employ attention mechanisms to enhance cross-modal understanding and manipulation. However, their efficiency is hampered when synthesizing images due to pixel-by-pixel generation.

\subsubsection{Other Models}
Other models have also been explored. Flow-based models seek to learn the data's probability density function via invertible transformations~\cite{Dinh2016DensityNVP}, allowing for data generation by sampling unobserved, realistic data points. This method has been applied to cross-modal translation, though it does not match the performance of other models~\cite{Wei2022Flow-basedImage, Sanghi2021CLIP-Forge:Generation}. Implicit field models, trained as classifiers, have been used for shape synthesis~\cite{Chen2018LearningModeling, Liu2019LearningSupervision, Park2019DeepSDF:Representation}. These models use embeddings from shape encoders and point coordinates to assign a binary value to each point, indicating its location relative to the shape, thus inferring the shape by surface point sampling.

\subsubsection{Model Evaluation}
For cross-modal translation, synthesis quality is evaluated via discriminability and diversity. Metrics like the inception score~\cite{Salimans2016ImprovedGANs} and Frechet inception distance~\cite{Heusel2017GANsEquilibrium} gauge image realism, showing high correlation with human judgment. Multi-scale structural similarity~\cite{Odena2016ConditionalGANs} assesses image diversity, while R-precision~\cite{Xu2018AttnGAN:Networks} and visual-semantic similarity\cite{Zhang2018PhotographicNetwork} evaluate the consistency between generated images and text prompts. For text-guided image editing, evaluations consider the correspondence between synthesized images and text prompts and the preservation of original images' irrelevant attributes. While cosine similarity, peak signal-to-noise ratio, and structural similarity only assess image-text correspondence, manipulative precision~\cite{Li2019ManiGAN:Manipulation} considers both correspondence and irrelevant attribute preservation. 3D shape synthesis quality is primarily gauged by similarity metrics. For point cloud-represented shapes, Chamfer distance~\cite{Achlioptas2017LearningClouds}, earth mover distance~\cite{Achlioptas2017LearningClouds}, and latent feature comparison~\cite{Shu20193DConvolutions} are used. Surface-represented shapes, like meshes, are assessed using the light field descriptor~\cite{Chen2018LearningModeling} and minimum matching distance~\cite{Ibing20213DFunctions}. Intersection over union and F-score measure reconstruction accuracy of voxel-represented shapes~\cite{Sbrolli2022IC3D:Generation}. Additionally, human evaluation is also used to determine the realism and consistency of the generated samples with the input guidance.

Beyond translation, cross-modal correlation and alignment captured by MMML models also facilitate inter-modal learning, a topic we will delve into in the following section.

\subsection{Co-learning}
Co-learning, as defined in~\cite{Baltrusaitis2019multimodalTaxonomy}, involves knowledge transfer from data-rich to data-scarce modalities during the training of MMML models. Typically, only the data-scarce modalities are available during testing.

\subsubsection{Transfer Learning}
\begin{figure*}[!ht]
    \centering
    \includegraphics[width=16cm]{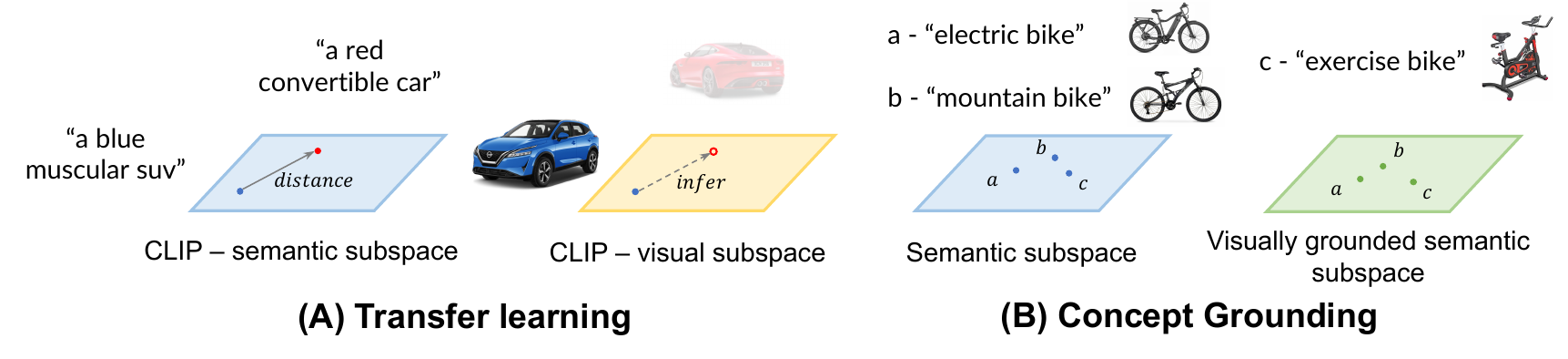}
    \caption{ross-modal co-learning: A) Given ``a blue muscular suv" and ``a red convertible car" in the semantic domain, with only the SUV visually represented, the convertible can be inferred by applying the semantic distance to the visual domain. B) Semantically, a ``mountain bile" (b) is closer to an ``exercise car" (c) than an ``electric car" (a). When visual information is introduced, b is closer to a, reflecting human understanding.}
    \label{fig:co-learning}
\end{figure*}

Coordinated representations in MMML enable knowledge transfer across modalities, permitting zero-shot prediction in data-scarce domains~\cite{Socher2013Zero-ShotTransfer, Frome2013DeViSE:Model, Tsai2018LearningRepresentations, BaPredictingDescriptions, Reed2016LearningDescriptions}. This can be seen in Figure~\ref{fig:co-learning}-A, where textual differences between design instances inform the unknown in the image domain~\cite{Socher2013Zero-ShotTransfer, Frome2013DeViSE:Model}. This knowledge transfer also aids in tasks like machine translation or document transliteration, using a pivot modality as a bridge between unassociated modalities~\cite{Rajendran2015BridgeLearning, Nakov2009ImprovedLanguages}. Additionally, semi-supervised MMML models can apply knowledge from labeled data to unlabeled data across modalities in situations with limited parallel labeled data~\cite{Hendricks2015DeepData, Socher2010ConnectingCorpora}. Furthermore, MMML models can yield improved unimodal representations even when only a single modality is available during testing~\cite{Desai2020VirTex:Annotations, BulentSariyildiz2020LearningAnnotations}.

\subsubsection{Concept Grounding}
Concept grounding uses multiple modalities like vision and sound to strengthen semantic understanding, much like humans use sensorimotor experience and perceptual information to ground concepts. In ML, various models project visual and semantic concepts into a shared space to enhance semantic representations~\cite{Socher2014GroundedSentences, Feng2010VisualRepresentation}, as shown in the bike design example in Figure~\ref{fig:co-learning}-B. With non-parallel multi-modal data, separate semantic and visual representations can be combined to enrich semantic comprehension~\cite{Shutova2016BlackFeatures, Socher2014GroundedSentences, Bruni2012DistributionalTechnicolor}. For instance, Vis-W2V adapts the original word2vec embeddings by capturing visual semantic relatedness~\cite{Kottur2016VisualWord2VecScenes}, while ViCo extends GloVe embeddings to incorporate visual co-occurrence data from Visual Genome~\cite{Gupta2019ViCo:Co-occurrences}. Visual concept grounding, an extension of concept grounding, learns visual concepts through semantic and visual inputs. It uses natural language supervision, rather than human-generated labels, to improve data efficiency and the applicability of the visual embeddings~\cite{Mori1999Image-to-wordDividing, Quattoni2007LearningCaptions, Joulin2015LearningData, Li2016LearningData}. Some models leverage weak supervision, like Instagram hashtags and image captions, to learn image representations, offering high transferability due to the rich information in large multi-modal datasets~\cite{Mahajan2018ExploringPretraining, Desai2020VirTex:Annotations, BulentSariyildiz2020LearningAnnotations}. However, concept grounding only improves performance if the extra grounding information is relevant to the task~\cite{BulentSariyildiz2020LearningAnnotations, Kiela2015GroundingPerception}.

ICross-modal co-learning also aids data augmentation, leveraging parallel multi-modal data to co-train multiple predictors using individual modalities. This process facilitates discovering additional labeled training samples, making use of inexpensive unlabeled data to augment smaller labeled datasets~\cite{Blum1998CombiningCo-training, Levin2003UnsupervisedCo-training}. It can also detect unreliable training samples~\cite{Christoudias2012Multi-ViewDisagreement}, though it risks data bias and overfitting~\cite{Baltrusaitis2019multimodalTaxonomy}. Furthermore, concept grounding using multi-modal data enables more accurate representations of design spaces, as illustrated in Figure~\ref{fig:co-learning}-B. This helps in design evaluation and informs optimization decisions. For instance, ``electric bike" might appear as the most novel design semantically, while ``exercise bike" ranks highest in novelty when grounding the semantic concepts visually, aligning with human perception. Building on these fundamentals, we will now review MMML applications that hold potential for engineering design.

\section{Applications of Multi-modal Machine Learning}
This section discusses three major categories of MMML applications: 1) cross-modal synthesis; 2) multi-modal prediction; and 3) cross-modal reasoning. Since the approaches of cross-modal translation have been discussed, we will focus here on their applications to text, image, and shape syntheses. Multi-modal prediction examines how MMML can enhance classification and regression outcomes. Cross-modal information retrieval pertains to studies employing reasoning with knowledge from one modality to respond to queries or questions in another. Table~\ref{table:challenges} summarizes the foundational concepts involved in each application category.

\begin{table*}[!ht]
\caption{The foundational concepts involved in different MMML applications}
\small
\begin{center}
\label{table:challenges}
\begin{tabular}{l c c c c c}
\hline

\hline
 & \textbf{Representation} & \textbf{Alignment} & \textbf{Fusion} & \textbf{Translation} & \textbf{Co-learning} \\
\hline
\textbf{Cross-modal Synthesis} & \checkmark & \checkmark & \checkmark  & \checkmark & \checkmark\\
\hline
\textbf{Multi-modal Prediction} & \checkmark & \checkmark & \checkmark &  & \checkmark \\
\hline
\textbf{Cross-modal Information Retrieval} & \checkmark & \checkmark &  &  & \checkmark \\
\hline
\end{tabular}
\end{center}
\end{table*}


\subsection{Cross-modal Synthesis}
Cross-modal synthesis, a significant aspect of MMML, has been the subject of extensive research due to its complexity. It poses challenges in three areas: 1) understanding the source mode instance and its key elements; 2) generating the target mode instance accurately, comprehensively, and succinctly; and 3) evaluating the open-ended task output. This section reviews cross-modal synthesis models used for generating text, images, and shapes.

\subsubsection{Visual-to-Text Synthesis}
Visual-to-text synthesis, particularly image captioning, has been extensively researched. Early rule-based models have given way to DL approaches, which use image encoders to capture visual features and ARM decoders to generate corresponding captions. CNN- and transformer-based encoders have been employed to extract visual features~\cite{Girshick2015FastR-CNN, Wang2022End-to-EndCaptioning}, with ARM decoders, such as RNNs and transformers, generating captions as word sequences~\cite{Mao2015GenerationDescriptions, Vinyals2014ShowGenerator, Rohrbach2015TheDescription}. Despite their effectiveness, RNN-based models struggle with fine alignment between image regions and caption words~\cite{Mao2015GenerationDescriptions, Vinyals2014ShowGenerator, Rohrbach2015TheDescription}, leading to the use of visual attention~\cite{Xu2015AskAnswering, Anderson2017Bottom-UpAnswering} and guide vectors~\cite{LUO2021106873} for tighter visual-text correlation. Transformers, introduced in 2017, leverage masked self-attention and cross-modal attention for caption prediction and visual feature alignment~\cite{Cornia2022ExplainingAnalysis, Wang2022End-to-EndCaptioning}. Since the position encoding and attention mechanism of original transformers are arduous to encode the spatial relations between visual regions~\cite{Herdade2019ImageWords}, modifications are needed to better capture the spatial relations. Approaches include adding memory modules~\cite{Huang2019AttentionCaptioning, He2020ImageTransformer}, incorporating spatial attention~\cite{Herdade2019ImageWords}, or using dual-stream transformers with a gated bilateral controller~\cite{Li2019EntangledCaptioning}. Alternatively, 1D CNNs have also been used as decoders~\cite{Aneja2017ConvolutionalCaptioning, Deshpande2018FastPart-of-Speech}.

In engineering design, visual-to-text synthesis can automate natural language descriptions of visually represented designs. With the rise of online platforms like Pinterest and Fusion 360 Gallery, design ideas are frequently shared through sketches, images, 3D models, and renderings, often with limited or no textual descriptions. This absence can hinder efficient design retrieval and inspiration search, and impede access to multi-modal data necessary for effective MMML model training. Visual-to-text synthesis can be applied to mitigate this issue.

\subsubsection{Cross-modal image Synthesis}
\begin{figure}[!ht]
    \centering
    \includegraphics[width=8cm]{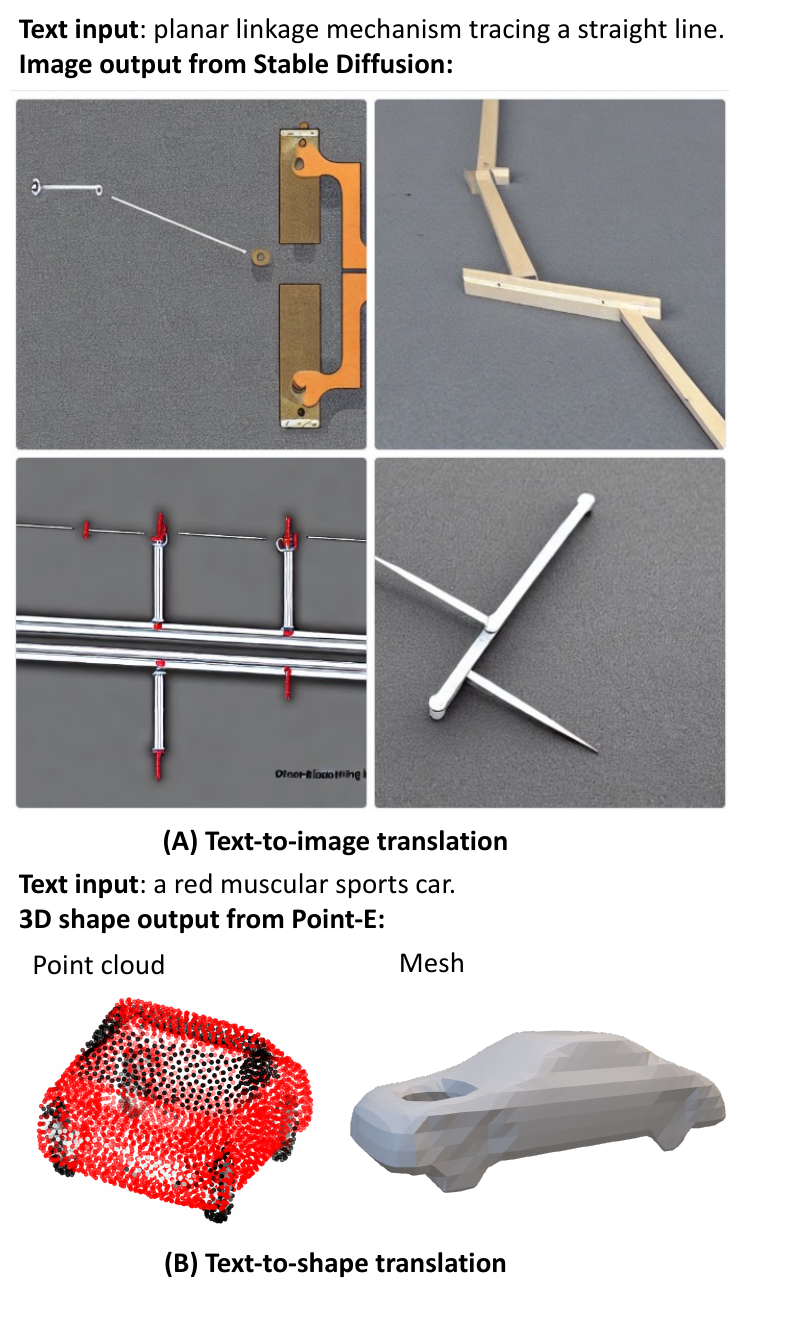}
    \caption{Cross-modal synthesis: A) Planar linkage mechanisms from stable diffusion~\cite{Rombach2021High-ResolutionModels} do not adhere to the ``tracing a straight line" requirement. B) The "muscular sports car" from Point-E~\cite{Nichol2022Point-E:Prompts} displays an unintended hood hole and lacks distinctive sports car features.}
    \label{fig:example}
\end{figure}

Text-to-2D synthesis, a task that generates 2D visuals based on input prompts, is a compelling research area in MMML. Recent advancements have employed various CGANs, DDMs, and ARMs to produce high-quality 2D visuals from abstract inputs like text, semantic maps, or depth maps. This section showcases their capabilities, exemplified by text-to-image translation from stable diffusion (Figure~\ref{fig:example}-A).

CGANs were seminal in text-guided image synthesis. Early versions yielded low-resolution images, but advancements came with stacked CGANs, such as StackGAN~\cite{Zhang2016StackGAN:Networks}, which generated preliminary shapes and colors in its first stage, and high-resolution images in the second. Further enhancements included DM-GAN's dynamic memory module~\cite{Zhu2019DM-GAN:Synthesis} and StackGAN++'s multi-branch structure~\cite{Zhang2019StackGAN++:Networks}. On this basis, AttnGAN~\cite{Xu2018AttnGAN:Networks} used a visual attention mechanism for stronger text-image alignment. Other variants improved image resolution or alignment differently~\cite{Zhang2018PhotographicNetwork, Li2019ManiGAN:Manipulation, Li2019ControllableGeneration}. Some models used both conditional and unconditional losses to evaluate image authenticity and text match simultaneously~\cite{Zhang2019StackGAN++:Networks, Xu2018AttnGAN:Networks, Zhu2019DM-GAN:Synthesis, Zhang2018PhotographicNetwork, Dash2017TAC-GANNetworkb}. Others considered semantic relevance between text and images~\cite{Cha2018AdversarialSynthesis} or employed image-to-text reconstruction loss~\cite{Qiao2019MirrorGAN:Redescription}.

A subset of CGANs synthesizes spatially controlled or complex images conditioned on concept layouts. Reed et al.'s model generates images with one spatially located object, using a semantic map to condition a stacked CGAN~\cite{Reed2016LearningDraw}. Models like Layout2Im extend to multi-object synthesis, combining objects into one semantic map~\cite{Zhao2018ImageLayout}. Some models use scene graphs~\cite{Johnson2018ImageGraphs} or text descriptions~\cite{Hong2018InferringSynthesis, Li2019Object-drivenTraining} converted into semantic maps to guide CGANs. Instead of using semantic maps, a dual-path model is conditioned on layout and class labels to create backgrounds and objects separately~\cite{Hinz2019GeneratingLocations}. Though stacked CGANs still generate low-resolution images due to computational constraints, DF-GAN uses a one-stage process with a text-image fusion block for deeper semantic-visual feature fusion~\cite{Tao2020DF-GAN:Synthesis}. A few models, adapted from StyleGAN~\cite{Karras2018ANetworks} for cross-modal image synthesis, use a style-based generator for style control and a revised matching network to align the generated images with input prompts via CLIP embeddings~\cite{Patashnik2021StyleCLIP:Imagery, Gal2021StyleGAN-NADA:Generators, Chefer2021Image-BasedTransfer}.

The adaption of transformer-based ARMs for image synthesis is facilitated by quantized image representations, akin to a codebook rich in context, often acquired via a discrete CNN-based VAE~\cite{vandenOordDeepMind2017NeuralLearning}. Models like VQVAE~\cite{vandenOordDeepMind2017NeuralLearning} and DALL-E~\cite{Ramesh2021Zero-ShotGeneration} use ARMs for text-guided image manipulation via a two-stage process: the initial stage involves training the discrete VAE and generating quantized image representations, followed by an ARM decoder training stage that conditioned on text for image generation. In VQGAN, a quantized encoder embeds visual prompts into vectors, which the GAN model, equipped with an ARM generator, uses to condition image synthesis. Several adaptations of VQVAE and VQGAN have been created for text or visual-guided image manipulation~\cite{Crowson2022VQGAN-CLIP:Guidance, Yu2022ScalingGeneration, Ding2021CogView:Transformers, Yu2021Vector-quantizedVQGAN}. These models blend the local interaction learning prowess of CNNs with the transformers' ability for handling long-range interactions and learning fine-grained cross-modal controls, enabling high-resolution image synthesis.

Recently, conditional DDMs have gained prominence for image manipulation. Models like GLIDE utilize this approach for text-guided image synthesis, often relying on classifier-free guidance to condition the denoising process and achieving high photo realism and semantic-visual correspondence~\cite{Nichol2021GLIDE:Models}. Stable Diffusion uses a two-stage training process to first create a 2D latent image space through a VAE and then train a latent diffusion model within this space, boosting computational efficiency~\cite{Rombach2021High-ResolutionModels}. It also exploits a cross-modal attention mechanism to further refine image-input alignment. Imagen enhances image quality by stacking multiple DDMs~\cite{Saharia2022PhotorealisticUnderstanding}. Alternative conditioning approaches have also been explored. For instance, DiffusionCLIP utilizes a directional CLIP loss for multi-attribute manipulation~\cite{Kim2021DiffusionCLIP:Manipulation}, while DALL-E 2 follows the embedding-based approach to fuse prompt embeddings with sampled noise as input to condition a stacked denoising module~\cite{Ramesh2022HierarchicalLatents}.

Unlike conditional generative models, CLIPDraw employs gradient optimization for text-guided drawing synthesis by maximizing text-drawing CLIP similarity~\cite{Frans2021CLIPDraw:Encoders}.

Cross-modal 2D visual synthesis holds great promise for engineering design. Traditionally, design ideation relies on free-hand sketching and drawing, disqualifying many people without professional training as effective designers. For trained designers, it is time-consuming to manually sketch or draw designs. Conversely, expressing design ideas through natural language or simple semantic layouts is easier and faster. Then, the models reviewed above can convert abstract expressions to 2D design visuals automatically. AI-assisted design generation not only eases professional designers' workload but also democratizes design practice, potentially boosting large-scale design customization. This approach can also potentially enhance design creativity. Visualizing design ideas involves re-organizing designers' usable knowledge, which is open-ended but limited by their knowledge bases. Cross-modal synthesis models are often trained on extensive multidisciplinary datasets, resulting in broader knowledge bases underlying the models. This allows for interdisciplinary knowledge transfer during cross-modal design synthesis, sparking creative designs. Moreover, the AI-assisted approach yields more realistic visuals than the manual approach, facilitating early-stage inspections and evaluations. Practically, 2D design visuals are intermediate representations and are usually converted to 3D design representations for downstream tasks like detailed design, prototyping, and manufacturing.

\subsubsection{Cross-modal 3D Synthesis}
3D representations are critical in engineering design due to their faithful portrayal of shapes with high-level details, particularly in the middle to late design stages for design evaluation, optimization, and prototyping. Cross-modal shape synthesis, owing to the complexity of 3D representations, currently stands as a significant challenge. Cross-modal 3D synthesis is commonly conditioned on text and visual prompts. Given that shapes can be represented by structured voxels or unstructured meshes and point clouds, each representation enables the use of different DGMs. Figure~\ref{fig:example}-B exemplifies text-guided shape generation by Point-E~\cite{Nichol2022Point-E:Prompts}.

Multi-modal VAEs initially garnered attention for 2D-to-3D synthesis. Leveraging the success of CNNs in 2D synthesis, models like 3D-R2N2 use 3D CNNs to synthesize voxel-represented shapes, transforming one or more 2D views into 3D shapes~\cite{Choy20163D-R2N2:Reconstruction}. However, due to high computational costs, voxel representations often suffer from low resolution. Mesh R-CNN mitigates this by coupling a mask R-CNN 2D perception module with a 3D CNN voxel decoder to create coarse voxel representations, which are then converted to meshes and refined by a GCN module~\cite{Gkioxari2019MeshR-CNN}. MeshMVS further enhances this by conditioning the refinement modules on depth images estimated from the input~\cite{Shrestha2020MeshMVS:Reconstruction}. Alternatively, models like PointOutNet use point cloud representation, pairing an image encoder with a decoder to predict point cloud coordinates~\cite{Fan2016AImage}. AtlasNet improves point cloud quality by mapping the points onto multiple deformed parametric surfaces~\cite{Groueix2018AtlasNet:Generation}. Lastly, TEVAE employs a mesh VAE and a 3D extrusion encoder, minimizing both mesh reconstruction loss and Euclidean loss between mesh and extrusion encodings for unconditional and sketch-guided 3D synthesis~\cite{Li2022AAutoencoder}.

While CGANs and ARMs are widely used for cross-modal 2D synthesis, their applications in the 3D domain are underexplored. In general, GANs struggle with high-resolution 3D synthesis due to resolution constraints~\cite{Wu2016LearningModeling, Khan2019UnsupervisedModeling}. To address this, some approaches represent 3D shapes as 2D data. Achlioptas et al. represented point coordinates as a matrix and trained VAEs and GANs on this 2D representation, enabling semantic editing and manipulation~\cite{Achlioptas2017LearningClouds}. Other studies map 3D shapes to 2D parameter domains, generate 2D samples using GANs, and convert them into 3D meshes~\cite{Maron2017ConvolutionalCovers, Ben-Hamu2018Multi-chartModeling}. Models like Rank3DGAN~\cite{Saquil2020Rank3DGAN:Attributes} and XDGAN~\cite{Alhaija2022XDGAN:Space} extend this to conditional settings, allowing semantic manipulation during mesh generation. ShaperCrafter applies ARMs to the 3D domain, learning a vector-quantized implicit 3D representation through a point VQVAE and conditioning the ARM decoder on BERT text embeddings for 3D synthesis~\cite{Fu2022ShapeCrafter:Model}.

DDMs have found success in cross-modal syntheses of 3D shapes primarily represented by point clouds. Luo et al. developed a DDPM, conditioning the denoising process on latent representations learned by a flow-based model for conditional 3D synthesis~\cite{Luo2021DiffusionGeneration}. Point-Voxel Diffusion marries DDMs with point-voxel shape representations, enabling unconditional or depth-guided 3D synthesis and shape completion~\cite{Zhou20213DDiffusion}. Similar to stable diffusion, LION uses an LDM for guided shape synthesis, encoding voxel and point cloud representations into two latent spaces using two point-voxel-CNN VAEs, and training two latent score-based DDMs in these spaces~\cite{Zeng2022LION:Generation}. LION, when conditioned on CLIP embeddings, can perform image- and text-guided 3D synthesis. Point-E~\cite{Nichol2022Point-E:Prompts} incorporates GLIDE~\cite{Nichol2021GLIDE:Models} with a DDM to convert text prompts to 2D images and then to 3D RGB point clouds, where the DDM is conditioned by fusing images' CLIP embeddings with input to the denoising process through a transformer encoder. There are also efforts to apply DDMs to 3D voxels~\cite{Sbrolli2022IC3D:Generation} and tetrahedral tessellation representations~\cite{Kalischek2022TetrahedralGeneration}.

Implicit field models, such as IM-NET~\cite{Chen2018LearningModeling} and DeepSDF~\cite{Park2019DeepSDF:Representation}, have made cross-modal 3D synthesis more accessible. These models, using a shape's embedding and point coordinates, identify if a point is inside or outside the shape. Multimodal VAEs comprising image or text encoders and implicit shape decoders have been developed for text- or image-guided 3D synthesis~\cite{Liu2019LearningSupervision, Alwala2022Pre-trainReconstruction, Liu2022ISS:Generation}. Ibing et al. employed a VAE consisting of a voxel encoder and an implicit 3D decoder to learn latent 3D space, and a GAN to generate new samples within the latent space, which can be conditioned on bounding boxes or class labels~\cite{Ibing20213DFunctions}. Similarly, CLIP-Forge learns a latent 3D space using a 2D rendering encoder and an implicit 3D decoder and then generates new samples in the latent space using a flow-based model conditioned on CLIP embeddings of text or image prompts~\cite{Sanghi2021CLIP-Forge:Generation}. Implicit field variational auto-decoders (VAD) have also been used to learn latent 3D spaces~\cite{Park2019DeepSDF:Representation}, within which 3D-LDM trains an LDM to generate new samples with text or image conditioning~\cite{Nam20223D-LDM:Models}. Alternatively, multi-modal VADs use two 2D VADs and a 3D implicit filed VAD to learn a cross-modal latent space applied to sketches, RGB views, and 3D shapes, allowing for 3D manipulation using sketches or RGB views~\cite{Cheng2022Cross-ModalManipulation}.

Other models focus on mesh-represented 3D shape manipulation. Pixel2Mesh progressively deforms an initial shape (e.g., an ellipsoid) using GCNs to align synthesized meshes with perceptual features of visual prompts~\cite{Wang2018Pixel2Mesh:Images}. Similarly, Text2Mesh~\cite{Michel2021Text2Mesh:Meshes} and ClipMatrix~\cite{Jetchev2021ClipMatrix:Meshes} employ multi-layer perceptions (MLPs) to predict mesh vertex shifts and color nuances for text-guided texture manipulation, maximizing CLIP similarities between 2D renderings of meshes and text prompts.

Cross-modal 3D synthesis in engineering design is a challenging and resource-intensive task for both humans and MMML models. The emergence of large pre-trained multi-modal representations like CLIP and CISP makes this task increasingly feasible and lessens the need for scarce parallel data. Such synthesis not only accelerates early-stage design - from concept generation to detailed design - but also enriches the pool of design ideas by allowing for easy visualization, concretization, and modeling. By converting text prompts, images, and sketches into high-fidelity shapes automatically, this technology can ease the human workload and expedite the design process. It also broadens design exploration. With released constraints of time and effort in creating 3D designs, designers can now bring more concepts to later stages for close inspection. These 3D shapes allow for detailed qualitative and quantitative evaluations, from visual appeal to spatial examination and to performance analysis, and further encourage optimization and ideation of new designs. Moreover, this technology enhances knowledge transfer and reuse in design. Unlike human designers limited by their knowledge and imagination, AI models tap into extensive multi-modal and multi-disciplinary datasets and can exploit the rich knowledge bases underlying them, potentially yielding more diverse and innovative designs than human designers.

Therefore, cross-modal 3D synthesis holds the potential to improve engineering design outcomes by expediting iterations, promoting exploration, and enhancing creativity. Typically, designers move from abstract descriptions to 3D models via 2D sketches as an intermediate step. Text-to-shape synthesis may enable designers to skip the sketching step, simplifying and accelerating the design process. However, this task remains challenging due to low precision and fidelity in generated shapes, as shown in Figure~\ref{fig:example}-B. This is attributed to the difficulty in expressing 3D features in rich detail using language and the scarcity of text-shape datasets for model training. As it is relatively easier for AI models to transfer visual features from 2D to 3D, a two-step process of text-to-image and image-to-shape synthesis can help manage this task. While the text-to-image step is currently more refined, the latter still needs further development.

\subsection{Multi-modal Prediction}
Multi-modal prediction techniques, like classification and regression, leverage various data modalities to enhance comprehension comprehensiveness and prediction accuracy. Leading models typically encode multi-modal data using joint representation and then proceed to predictions. The key differentiation among models lies in their encoding strategies and methods.

\subsubsection{Multi-modal Classification}
Classification is a key task in MMML with various applications across domains. For instance, the HyCon model employs hybrid contrastive learning to capture both intra-modal and inter-modal interactions for multi-modal sentiment analysis, learning from audio, visual, and text data~\cite{hycon}. The deep multi-modal design evaluation model predicts overall and category-specific sentiments using design images and view hierarchy, featuring a bidirectional encoder and a self-attention-based fusion model~\cite{Yuan2022LeveragingModel}. TechDoc, predicting patent classes, first combines patent text and images through concatenation and then integrates the bi-modal representation with patent associations through a GNN, which outperforms unimodal models~\cite{Jiang2022DeepClassification}. For fake news detection, FND-CLIP fuses text and image information by weighing visual and textual features based on cross-modal similarities and then employs a cross-modal attention mechanism to integrate the unimodal and fused features for final classification~\cite{fnd-clip}. Novel modalities like chemical substructures, targets, pathways, and enzymes are used in a model called DDIMDL to learn drug-drug interactions~\cite{ddimdl}. Thus, MMML's diverse applications utilize different modalities and models, demonstrating its flexibility and effectiveness.

\subsubsection{Multi-modal Regression}
MMML can greatly enhance the precision and efficiency of complex regression tasks. Various works concentrate on devising superior fusion schemes for multi-modal data, aiding downstream regression tasks. For instance, Song et al. proposed an MMML model, fusing visual and semantic embeddings via a symmetric cross-modal attention mechanism, to predict design metrics using sketches and text descriptions~\cite{Song2023ATTENTION-ENHANCEDEVALUATIONS}. Similarly, Yuan et al. used an attention mechanism to fuse image and text descriptions for predicting customer evaluation scores~\cite{Yuan2022LeveragingModel}. Other works incorporate additional modalities for improvement. DeepTake, a novel architecture for driver takeover prediction in autonomous vehicles, stands out~\cite{deeptake}. It predicts the intention, timing, and quality of human takeover based on various data types such as vehicle states, non-driving tasks, biometrics, and pre-driving survey. This work is challenging and novel due to the increased number of modalities involved.

Multi-modal prediction also facilitates automatic design synthesis. Synthesized designs must be visually realistic and comply with engineering standards, user needs, and market preferences. While designs are often multi-modal, most existing surrogate models used for design evaluation rely on unimodal data, limiting their accuracy and scope. However, multi-modal prediction techniques like those proposed by Song et al.~\cite{Song2023ATTENTION-ENHANCEDEVALUATIONS} and Yuan et al.~\cite{Yuan2022LeveragingModel} effectively evaluate designs using images and text descriptions, outperforming unimodal methods~\cite{10.1115/1.4056500}. Classification models handle categorical evaluations (e.g., design validity), and regression models deal with real-value attribute assessments (e.g., car drag coefficients, drone flying ranges). Inputs can range from text descriptions and sketches for conceptual designs, to shape and specification pairs for detailed designs, or image, specification, and text trios for evaluating user and market preferences on e-commerce sites.

\subsection{Cross-modal Information Retrieval}
Traditional information retrieval (IR) primarily uses natural language processing to align queries with textual content. Cross-modal knowledge extraction, however, broadens the scope by identifying correlations or alignments across different modalities.

Cross-modal IR involves finding instances in a target mode that closely align with source mode queries. Cross-modal IR can be conducted in either unimodal or multi-modal approaches, based on the representations used. Unimodal retrieval, focused on source mode instance representations, first retrieves instances nearest to the search query using similarity metrics~\cite{Yagcioglu2015ACaptioning, Ordonez2011Im2Text:Photographs} or k-nearest neighbor models~\cite{Devlin2015LanguageWorks, Kwon2022EnablingLearning}. The corresponding target mode instances are then returned as candidate retrievals. However, this approach applies only to parallel data, and high source mode similarities do not necessarily yield good cross-modal results, though careful metric design can mitigate this~\cite{Yagcioglu2015ACaptioning, Ordonez2011Im2Text:Photographs}. Conversely, multi-modal retrieval projects diverse modalities into a shared multi-modal space, where relevant instances are located based on query-candidate similarities. This approach, leveraging various multi-modal representation spaces, yields more expressive representations that reflect all modalities and leads to improved retrieval~\cite{Farhadi2010EveryImages, Socher2014GroundedSentences, Xu2015JointlyFramework, Hodosh2013FramingMetrics, Karpathy2014DeepMapping, Cao2016DeepRetrieval}. Moreover, a common multi-modal representation space enables bidirectional retrieval. Both unimodal and multi-modal methods strive for effective representations for high-quality cross-modal retrieval.

Cross-modal IR in engineering design aids in design knowledge gathering and design analogy and inspiration search. Many design repositories, like Pinterest and Fusion 360 Gallery, offer rich design precedents, often with minimal or no descriptions. This absence of textual information can hinder effective knowledge reuse. However, with annotated parallel datasets, we can train cross-modal IR models to correlate textual queries with 2D or 3D design representations. Transferring this learned knowledge to unimodal repositories helps overcome the lack of text, potentially enhancing the efficiency and accuracy of design knowledge retrieval.

\textcolor{black} {\section{Challenges and Opportunities of Multi-modal Machine Learning in Engineering Design}}
\textcolor{black} {The applications reviewed reveal the transformative potential of MMML for engineering design. By incorporating diverse data sources—2D visuals, text, 3D shapes, and more—MMML models can gain a comprehensive, multi-faceted understanding of design problems, enabling more informed decision-making. MMML can automate parts of the design process, significantly reducing the time and effort needed for design synthesis and enabling broader design exploration. MMML also accelerates design evaluation by providing faster, cost-effective, and increasingly precise assessments compared to traditional simulations and experiments. These applications can promote design customization and personalization by optimizing designs according to individual preferences and needs. With its potential to boost creativity, improve product quality, and enhance efficiency, MMML is becoming a valuable tool in the design field.}

\textcolor{black} {MMML can also enhance human-AI collaboration in engineering design. Design representations span various data modes, each differing in expressivity, abstraction, and complexity. Humans, with their strong cognitive abilities, excel at guiding exploration processes and decision-making at abstract levels, such as formulating and describing abstract design ideas—a task AI often struggles with. AI, conversely, with its computational prowess, excels at handling more expressive, complex information, such as crafting or evaluating intricate design representations. MMML allows both humans and AI to focus on tasks they are best suited for, leveraging their unique strengths to enhance overall performance.}

\textcolor{black} {However, applying MMML in engineering design comes with unique challenges that can impact its performance and effectiveness. Firstly, design spaces are more intricate compared to many other data domains. Design representations encompass structural, functional, and behavioral information~\cite{Gero2019TheModel}, which is often embodied in different modalities. While visual representations provide a direct depiction of structural and spatial relationships between design components, text descriptions articulate the functions and behavior of a design more explicitly. Second, design representations evolve across different design stages. In the early stages, abstract representations enable designers to effectively communicate their initial ideas and quickly modify and refine their designs, fostering creativity and encouraging broader exploration. In the later stages, high-fidelity representations offer a comprehensive, detailed view of designs, enabling better communication and coordination, simulation-based evaluation and optimization, and manufacturing preparation. The diversity and complexity of design spaces pose unique challenges and opportunities in applying MMML to engineering design, which we will discuss in this section.}

\textcolor{black} {\subsection{A Need for Large Multi-modal Design Datasets}}
\textcolor{black} {The collection of parallel multi-modal data is challenging, leading to a scarcity of multi-modal datasets for MMML. There are a few large parallel text-image datasets like MS-COCO~\cite{Lin2014MicrosoftContext}, Visual Genome~\cite{Krishna2016VisualAnnotations}, YFCC100M~\cite{Thomee2015YFCC100M:Research}, JFT-300M~\cite{Sun2017RevisitingEra}, and AVA~\cite{Murray2012AVA:Analysis} available for tasks like semantic-visual representation learning, image captioning, and text-to-image synthesis. Text2Shape~\cite{Chen2018Text2Shape:Embeddings} and Text2Shape++~\cite{Fu2022ShapeCrafter:Model} provide parallel text-shape datasets for semantic-geometric representation learning and text-to-shape synthesis. The scope of these generic datasets is limited to specific object classes and lacks design-specific information required for engineering design applications. Large accessible design datasets like Pinterest, ShapeNet, and Fusion 360 Gallery primarily contain single-modality design instances, making them unsuitable for MMML which requires aligned multi-modal data. The scarcity of multi-modal design data, coupled with the resource-intensive nature of design data labeling, often through simulations or experiments, impedes scaling design labeling processes and the creation of expansive multi-modal design datasets.}

\textcolor{black} {The dearth of high-quality, multi-modal design data hampers the training of MMML models for engineering design, hindering them from fully grasping diverse and intricate design spaces~\cite{Song2023ATTENTION-ENHANCEDEVALUATIONS}. Although transfer learning can somewhat mitigate this problem~\cite{Jahan2021ParkinsonsLearning}, general ML datasets may lack design-specific insights. This necessitates the creation of extensive multi-modal, high-quality design datasets to effectively train MMML models and facilitate knowledge transfer for various design tasks. The community should collaboratively construct and maintain expansive design datasets with high-quality labels. This would entail collecting and storing aligned multi-modal design data, labeling datasets with design-related attributes, and if available, providing pre-trained embeddings or latent representations, and specifying associated design context. Such efforts would provide significant benefits to the engineering design community.}

\textcolor{black} {\subsection{A Need for Data-driven Design Evaluation Models}}
\textcolor{black} {In general MMML, multi-modal information fusion affects the learning of not only intra-modality features but also inter-modality interactions, which may result in more informative multi-modal representations, enhancing downstream multi-modal prediction~\cite{Perez-Rua2019MFAS:Search}. Unlike standard prediction tasks, data-driven design evaluation relies on comprehending designs from three aspects: 1) a comprehensive understanding of intricate structural, functional, and behavioral design features conveyed by various modalities, 2) effective information fusion for capturing inter-modality interactions, and 3) complex reasoning with multi-modal features and inter-modality interactions~\cite{Song2023ATTENTION-ENHANCEDEVALUATIONS}. The success of these aspects requires more effective information fusion techniques to capture, fuse, and reason with various, complementary design features, fostering data-driven design evaluation.}

\textcolor{black} {Data-driven design evaluation is an important step toward applying cross-modal synthesis to broaden and expedite design exploration and optimization. General cross-modal synthesis models are typically evaluated on synthesis fidelity and cross-modal alignment~\cite{Zhang2018PhotographicNetwork, Salimans2016ImprovedGANs, Heusel2017GANsEquilibrium}. For engineering design applications, the usefulness of these models further hinges on whether the synthesized designs are valid, innovative, functional, and high-performing~\cite{Regenwetter2023BeyondDesign}. However, these design-specific evaluations are unaddressed in other domains, and we need to develop appropriate metrics and effective approaches for this. Traditional approaches to design evaluations typically rely on time-intensive, non-gradient-based simulations or experiments. This disables real-time design evaluation during the training of cross-modal synthesis models. Accordingly, data-driven approaches are better options to evaluate the quality of synthesized designs. With the availability of labeled design datasets, surrogate models can be developed for design quality evaluations during design synthesis. Since large labeled design datasets are not easily accessible, increasing data efficiency during training such models can augment their utility.}

\textcolor{black} {\subsection{A Need for More Effective Representation Learning Techniques}}
\textcolor{black} {Engineering design demands more effective multi-modal representations to capture the complexity of design spaces. MMML models in engineering design must seamlessly analyze and fuse information from various modalities to effectively model these design spaces and gain a comprehensive understanding of them. Effective techniques are needed to encode different design representations. While text and image data are predominant in ML due to their wide availability~\cite{Baltrusaitis2019multimodalTaxonomy}, 2D design visuals (e.g., sketches and design diagrams) and 3D shapes, which are common in engineering design, are less explored. Despite CNN-based models applicable to 2D visuals, these models struggle with comprehending design sketches or diagrams due to their sparsity~\cite{Xu2022DeepSurvey}. Moreover, while 3D CNNs and GNNs can handle 3D data, they primarily capture local features or coarse global features, missing fine-grained ones~\cite{Ghadai2019Multi-levelFeatures}. Given the importance of 3D representations in design evaluations and optimizations, there is a demand for effective 3D shape encoding techniques. This challenge presents an opportunity for design researchers to create specialized ML architectures for common design modalities.}

\textcolor{black} {Robust methods to manage ambiguity and inconsistency in design representations are also essential. Design sketches and drawings often display personal styles and contain casual annotations, making similar ideas appear in varied styles and detail levels~\cite{Song2023ATTENTION-ENHANCEDEVALUATIONS}. This variability increases when representing complex products and occasionally, certain modalities may be absent or unobtainable. Current ML models struggle to differentiate conceptual variations from representation discrepancies, account for missing modalities, and identify levels of abstraction and detail. To address these issues, the design community must develop MMML models that are robust to representation inconsistency, missing modalities, and varying detail levels. As Humans can more readily perceive and adjust representation discrepancies, human-AI hybrid teaming, enabling humans and multi-modal AI to complement each other, can be a promising strategy to tackle these complexities.}

\textcolor{black} {\subsection{A Need for More Effective Information Aligning Techniques}}
\textcolor{black} {Various existing attention mechanisms can effectively capture implicit feature alignment across modalities, supporting simple knowledge reasoning for cross-modal synthesis and co-learning~\cite{Mansimov2015GeneratingAttention, Xu2018AttnGAN:Networks, Li2019Object-drivenTraining}. However, this is insufficient for applications in engineering design. Understanding engineering designs necessitates capturing complex interrelations and interactions~\footnote{As an illustration, rotors and fixed wings, despite their structural differences, both provide lift for aircraft, albeit via different mechanisms.} among different components in terms of function, structure, and behavior ideally in an explicit way. Although a few datasets~\cite{Kong2014WhatCoreference, Plummer2015Flickr30kModels, Mao2015GenerationDescriptions} are annotated to support explicit alignment learning, they are neither from the engineering domain nor tailored for design applications. Therefore, the need arises for well-annotated design datasets to learn fine-grained cross-modal alignment for complicated design knowledge reasoning.}

\textcolor{black} {Moreover, most attention mechanisms primarily excel in learning implicit semantic-visual alignment due to the abundance of textual and visual data. As mentioned above, engineering design often involves sketches and 3D shapes, beyond text and images. The effectiveness of the existing attention mechanisms for aligning sketch and 3D shape features has not been validated concretely. This necessitates sophisticated adaptations of existing mechanisms and the development of new ones to effectively align design-specific features.}

\textcolor{black} {\subsection{A Need for Pre-trained and Generalizable Multi-modal Representations}
In MMML, pre-trained multi-modal representations facilitate cross-modal co-learning, facilitating knowledge transfer from sample-rich modalities to sample-scarce modalities and mitigating data scarcity issues common in MMML for engineering design~\cite{Radford2021LearningSupervision, Sbrolli2022IC3D:Generation}. The advancement of pre-trained semantic-visual representation models like CLIP has greatly promoted the development of cross-modal synthesis models~\cite{Radford2021LearningSupervision}. However, as the pre-trained representations were not specifically trained on design data, the underlying design knowledge remains limited. Specifically, these models struggle to accurately reflect qualitative and quantitative design descriptions, requirements, and specifications, which can lead to invalid or suboptimal outcomes in design evaluation and synthesis. For instance, in Figure~\ref{fig:example}-A, the pre-trained model misinterprets the design requirement ``planar linkage mechanism \emph{tracing a straight line}," resulting in unsatisfactory linkage mechanisms. To bolster engineering design applications, the design community should either enhance existing pre-trained models with more domain knowledge or develop new models trained on comprehensive design datasets to glean more domain-specific understanding during multi-modal representation learning.}

\textcolor{black} {\subsection{A Need for Model Scalability and Interpretability}}
\textcolor{black} {Beyond text and visual processing, interpreting, evaluating, and optimizing engineering designs also involve comprehending additional data modes like olfactory, haptic, auditory, emotional, and ergonomic modes~\cite{Wu2022ResearchCOVID-19}. The insufficient exploration of these data modes necessitates effective models for their processing and inclusion into MMML. To enhance model comprehensiveness and effectiveness, specific efforts need to be made to expand MMML architectures and accommodate a wider range of data modes for engineering design applications.}

\textcolor{black} {Additionally, the interpretability of MMML models, the ability to understand and explain their decision-making process, is important but lacking. In general ML, a model's interpretability not only builds trust in its predictions but also helps avoid or uncover unexpected biases~\cite{Linardatos2021ExplainableMethods, BarredoArrieta2020ExplainableAI}. Interpretability in MMML can further inform users about the importance of each involved modality and the specific cross-modal interactions that affect decision-making, potentially enabling them to learn ``implicit" knowledge from MMML models. Given the ``black box" nature of deep MMML models, it is challenging to improve their interpretability. A future direction is to balance MMML performance with interpretability, making MMML models more interactive, trustable, and effective for human designers.}

\section{Conclusion}
In conclusion, multi-modal machine learning (MMML) holds the potential to usher in a new era of engineering design by effectively capturing, integrating, and reasoning with design knowledge buried in various forms of design data. This paper provides a comprehensive overview of the fundamental concepts in MMML, including multi-modal information representation, fusion, alignment, translation, and co-learning. These evolving techniques underpin the cutting-edge applications of MMML. We also highlighted significant use cases relevant to engineering design, such as cross-model synthesis, multi-modal prediction, and cross-modal information retrieval. Despite its potential, MMML confronts unique challenges in the field of engineering design. This paper outlines the se research gaps and encourages efforts towards building large multi-modal design datasets, devising effective data-driven MMML techniques specific to design applications, and improving the scalability and interpretability of MMML models. Furthermore, rigorous empirical studies are essential to confirm the practical effectiveness of MMML in real-world engineering design applications.






%
% ...

\bibliographystyle{asmems4}

\bibliography{asme2e}

\begin{thebibliography}{100}

\bibitem{Bengio2012RepresentationPerspectives}
Bengio, Y., Courville, A., and Vincent, P., 2012,
\newblock ``{Representation Learning: A Review and New Perspectives},''
\newblock {\em IEEE Transactions on Pattern Analysis and Machine Intelligence,
  {\bf 35}}(8), 6, pp.~1798--1828.

\bibitem{10.1115/1.4039450}
Bhattacharjee, K.~S., Singh, H.~K., and Ray, T., 2018,
\newblock ``{Multiple Surrogate-Assisted Many-Objective Optimization for
  Computationally Expensive Engineering Design},''
\newblock {\em Journal of Mechanical Design, {\bf 140}}(5), 03, p.~051403.

\bibitem{Zhu2023BiologicallyTransformers}
Zhu, Q., Zhang, X., and Luo, J., 2023,
\newblock ``{Biologically Inspired Design Concept Generation Using Generative
  Pre-Trained Transformers},''
\newblock {\em Journal of Mechanical Design, {\bf 145}}(4), 4.

\bibitem{Zhu2023GenerativeGeneration}
Zhu, Q., and Luo, J., 2023,
\newblock ``{Generative Transformers for Design Concept Generation},''
\newblock {\em Journal of Computing and Information Science in Engineering,
  {\bf 23}}(4), 8, pp.~1--61.

\bibitem{Nobari2021PcDGAN:Design}
Nobari, A.~H., Chen, W., and Ahmed, F., 2021,
\newblock ``{PcDGAN: A Continuous Conditional Diverse Generative Adversarial
  Network For Inverse Design},''
\newblock In 27th ACM SIGKDD Conference on Knowledge Discovery {\&} Data, ACM,
  pp.~610--616.

\bibitem{LUO2021106873}
Luo, J., Sarica, S., and Wood, K.~L., 2021,
\newblock ``{Guiding data-driven design ideation by knowledge distance},''
\newblock {\em Knowledge-Based Systems, {\bf 218}}, 4, p.~106873.

\bibitem{Song2023ATTENTION-ENHANCEDEVALUATIONS}
Song, B., Associate, P., Miller, S., and Ahmed, F., 2023,
\newblock ``{ATTENTION-ENHANCED MULTIMODAL LEARNING FOR CONCEPTUAL DESIGN
  EVALUATIONS},''
\newblock {\em Journal of Mechanical Design}, 1, pp.~1--38.

\bibitem{10.1115/1.4049895}
Feng, Y., Li, M., Lou, S., Zheng, H., Gao, Y., and Tan, J., 2021,
\newblock ``{A Digital Twin-Driven Method for Product Performance Evaluation
  Based on Intelligent Psycho-Physiological Analysis},''
\newblock {\em Journal of Computing and Information Science in Engineering,
  {\bf 21}}(3), 02, p.~031002.

\bibitem{Nobari2021Range-GAN:Synthesis}
Nobari, A.~H., Chen, W., and Ahmed, F., 2021,
\newblock ``{Range-GAN: Range-Constrained Generative Adversarial Network for
  Conditioned Design Synthesis},''
\newblock {\em Proceedings of the ASME Design Engineering Technical Conference,
  {\bf 3B-2021}}, 3.

\bibitem{Song2022AssessingPrediction}
Song, B., McComb, C., and Ahmed, F., 2022,
\newblock ``{Assessing Machine Learnability of Image and Graph Representations
  for Drone Performance Prediction},''
\newblock {\em Proceedings of the Design Society, {\bf 2}}, 5, pp.~1777--1786.

\bibitem{Gero1990DesignDesign}
Gero, J.~S., 1990,
\newblock ``{Design Prototypes: A Knowledge Representation Schema for
  Design},''
\newblock {\em AI Magazine, {\bf 11}}(4), 12, pp.~26--26.

\bibitem{Tseng2011HowTask}
Tseng, W.~S., and Ball, L.~J., 2011,
\newblock ``{How Uncertainty Helps Sketch Interpretation in a Design Task},''
\newblock {\em Design Creativity 2010}, pp.~257--264.

\bibitem{Haggman2015ConnectionsDesign}
H{\"{a}}ggman, A., Tsai, G., Elsen, C., Honda, T., and Yang, M.~C., 2015,
\newblock ``{Connections Between the Design Tool, Design Attributes, and User
  Preferences in Early Stage Design},''
\newblock {\em Journal of Mechanical Design, {\bf 137}}(7), 7.

\bibitem{Tsai2017HowCAD}
Tsai, G., and Yang, M.~C., 2017,
\newblock ``{How It Is Made Matters: Distinguishing Traits of Designs Created
  by Sketches, Prototypes, and CAD},''.

\bibitem{Purcell1998DrawingsPsychology}
Purcell, A.~T., and Gero, J.~S., 1998,
\newblock ``{Drawings and the design process: A review of protocol studies in
  design and other disciplines and related research in cognitive psychology},''
\newblock {\em Design Studies, {\bf 19}}(4), 10, pp.~389--430.

\bibitem{Ullman1990TheProcess}
Ullman, D.~G., Wood, S., and Craig, D., 1990,
\newblock ``{The importance of drawing in the mechanical design process},''
\newblock {\em Computers and Graphics, {\bf 14}}(2), pp.~263--274.

\bibitem{Chang2016EffectsAbilities}
Chang, Y.~S., Chien, Y.~H., Lin, H.~C., Chen, M.~Y., and Hsieh, H.~H., 2016,
\newblock ``{Effects of 3D CAD applications on the design creativity of
  students with different representational abilities},''
\newblock {\em Computers in Human Behavior, {\bf 65}}, 12, pp.~107--113.

\bibitem{Atilola2016TheTrees}
Atilola, O., Tomko, M., and Linsey, J.~S., 2016,
\newblock ``{The effects of representation on idea generation and design
  fixation: A study comparing sketches and function trees},''
\newblock {\em Design Studies, {\bf 42}}, 1, pp.~110--136.

\bibitem{Hannibal2016AnDesign}
Hannibal, C., Brown, A., and Knight, M., 2016,
\newblock ``{An Assessment of the Effectiveness of Sketch Representations in
  Early Stage Digital Design},''
\newblock {\em http://dx.doi.org/10.1260/1478077053739667, {\bf 3}}(1), 11,
  pp.~107--125.

\bibitem{Atilola2015RepresentingSketches}
Atilola, O., and Linsey, J., 2015,
\newblock ``{Representing analogies to influence fixation and creativity: A
  study comparing computer-aided design, photographs, and sketches},''
\newblock {\em Artificial Intelligence for Engineering Design, Analysis and
  Manufacturing: AIEDAM, {\bf 29}}(2), 4, pp.~161--171.

\bibitem{Reid2013ImpactJudgment}
Reid, T.~N., MacDonald, E.~F., and Du, P., 2013,
\newblock ``{Impact of Product Design Representation on Customer Judgment},''
\newblock {\em Journal of Mechanical Design, {\bf 135}}(9), 9.

\bibitem{Yang2005AOutcome}
Yang, M.~C., 2005,
\newblock ``{A study of prototypes, design activity, and design outcome},''
\newblock {\em Design Studies, {\bf 26}}(6), 11, pp.~649--669.

\bibitem{McKoy2020InfluenceGeneration}
McKoy, F.~L., Vargas-Hern{\'{a}}ndez, N., Summers, J.~D., and Shah, J.~J.,
  2020,
\newblock ``{Influence of Design Representation on Effectiveness of Idea
  Generation},''
\newblock {\em Proceedings of the ASME Design Engineering Technical Conference,
  {\bf 4}}, 11, pp.~39--48.

\bibitem{Grace2014Data-intensiveSurprise}
Grace, K., Maher, M.~L., Fisher, D., and Brady, K., 2014,
\newblock ``{Data-intensive evaluation of design creativity using novelty,
  value, and surprise},''
\newblock {\em International Journal of Design Creativity and Innovation, {\bf
  3}}(3-4), pp.~125--147.

\bibitem{Nomaguchi2019AssessingDesign}
Nomaguchi, Y., Kawahara, T., Shoda, K., and Fujita, K., 2019,
\newblock ``{Assessing Concept Novelty Potential with Lexical and
  Distributional Word Similarity for Innovative Design},''
\newblock {\em Proceedings of the Design Society: International Conference on
  Engineering Design, {\bf 1}}(1), pp.~1413--1422.

\bibitem{Wood2001ProductDevelopment}
Wood, K., and Otto, K., 2001,
\newblock {Product Design: Techniques in Reverse Engineering and New Product
  Development}.

\bibitem{Ulrich2000ProductDevelopment}
Ulrich, K.~T., and Eppinger, S.~D., 2000,
\newblock {\em {Product design and development}}
\newblock McGraw-Hill.

\bibitem{Fiorineschi2018IssuesAssessments}
Fiorineschi, L., Frillici, F.~S., and Rotini, F., 2018,
\newblock ``{Issues related to missing attributes in aposteriori novelty
  assessments},''
\newblock {\em Proceedings of International Design Conference, DESIGN, {\bf
  3}}, pp.~1067--1078.

\bibitem{Veisz2012Computer-aidedStudy}
Veisz, D., Namouz, E.~Z., Joshi, S., and Summers, J.~D., 2012,
\newblock ``{Computer-aided design versus sketching: An exploratory case
  study},''
\newblock {\em Artificial Intelligence for Engineering Design, Analysis and
  Manufacturing: AIEDAM, {\bf 26}}(3), 8, pp.~317--335.

\bibitem{BABAPOUR2014MediaEducation}
BABAPOUR, M., ORNAS, V. H.~A., REXFELT, O., and RAHE, U., 2014,
\newblock ``{Media and Representations in Product Design Education},''
\newblock In INTERNATIONAL CONFERENCE ON ENGINEERING AND PRODUCT DESIGN
  EDUCATION, E.~Bohemia, A.~Eger, W.~Eggink, A.~Kovacevic, B.~Parkinson, and
  W.~Wits, eds., pp.~42--47.

\bibitem{Baltrusaitis2019multimodalTaxonomy}
Baltrusaitis, T., Ahuja, C., and Morency, L.~P., 2019,
\newblock ``{Multimodal Machine Learning: A Survey and Taxonomy},''
\newblock {\em IEEE Transactions on Pattern Analysis and Machine Intelligence,
  {\bf 41}}(2), 2, pp.~423--443.

\bibitem{Zhang2019multimodalApplications}
Zhang, C., Yang, Z., He, X., and Deng, L., 2019,
\newblock ``{Multimodal Intelligence: Representation Learning, Information
  Fusion, and Applications},''
\newblock {\em IEEE Journal on Selected Topics in Signal Processing, {\bf
  14}}(3), 11, pp.~478--493.

\bibitem{Cui2022DeepReview}
Cui, C., Yang, H., Wang, Y., Zhao, S., Asad, Z., Coburn, L.~A., Wilson, K.~T.,
  Landman, B.~A., and Huo, Y., 2022,
\newblock ``{Deep Multi-modal Fusion of Image and Non-image Data in Disease
  Diagnosis and Prognosis: A Review},''.

\bibitem{zhenghui2022}
Li, X., Wang, Y., and Sha, Z., 2022,
\newblock ``{Deep-Learning Methods of Cross-Modal Tasks for Conceptual Design
  of Product Shapes: A Review},''
\newblock {\em Journal of Mechanical Design}, 12, pp.~1--31.

\bibitem{Dhariwal2021DiffusionSynthesis}
Dhariwal, P., and Nichol, A., 2021,
\newblock ``{Diffusion Models Beat GANs on Image Synthesis},''
\newblock {\em Advances in Neural Information Processing Systems, {\bf 11}}, 5,
  pp.~8780--8794.

\bibitem{Nichol2021GLIDE:Models}
Nichol, A., Dhariwal, P., Ramesh, A., Shyam, P., Mishkin, P., McGrew, B.,
  Sutskever, I., and Chen, M., 2021,
\newblock ``{GLIDE: Towards Photorealistic Image Generation and Editing with
  Text-Guided Diffusion Models},''.

\bibitem{Kim2021DiffusionCLIP:Manipulation}
Kim, G., Kwon, T., and Ye, J.~C., 2021,
\newblock ``{DiffusionCLIP: Text-Guided Diffusion Models for Robust Image
  Manipulation},''.

\bibitem{Frome2013DeViSE:Model}
Frome, A., Corrado, G.~S., Shlens, J., Bengio, S., Dean, J., Ranzato, M., and
  Mikolov, T., 2013,
\newblock ``{DeViSE: A Deep Visual-Semantic Embedding Model},''
\newblock {\em Advances in Neural Information Processing Systems, {\bf 26}}.

\bibitem{Rajendran2015BridgeLearning}
Rajendran, J., Khapra, M.~M., Chandar, S., and Ravindran, B., 2015,
\newblock ``{Bridge Correlational Neural Networks for Multilingual Multimodal
  Representation Learning},''
\newblock {\em 2016 Conference of the North American Chapter of the Association
  for Computational Linguistics: Human Language Technologies, NAACL HLT 2016 -
  Proceedings of the Conference}, 10, pp.~171--178.

\bibitem{Srivastava2012multimodalMachines}
Srivastava, N., and Salakhutdinov, R.~R., 2012,
\newblock ``{Multimodal Learning with Deep Boltzmann Machines},''
\newblock {\em Advances in Neural Information Processing Systems, {\bf 25}}.

\bibitem{DucTuan2021multimodalDetection}
Duc~Tuan, N.~M., and Quang Nhat~Minh, P., 2021,
\newblock ``{Multimodal Fusion with BERT and Attention Mechanism for Fake News
  Detection},''
\newblock {\em Proceedings - 2021 RIVF International Conference on Computing
  and Communication Technologies, RIVF 2021}, 4.

\bibitem{Song2022HEYAnd}
Song, B., Miller, S., and Ahmed, F., 2022,
\newblock ``Hey, ai! can you see what i see? multimodal transfer learning-based
  design metrics prediction for sketches with text descriptions,''
\newblock In International Design Engineering Technical Conferences and
  Computers and Information in Engineering Conference, Vol.~86267, American
  Society of Mechanical Engineers, p.~V006T06A017.

\bibitem{Yuan2022LeveragingModel}
Yuan, C., Marion, T., and Moghaddam, M., 2022,
\newblock ``{Leveraging End-User Data for Enhanced Design Concept Evaluation: A
  Multimodal Deep Regression Model},''
\newblock {\em Journal of Mechanical Design, {\bf 144}}(2), 2, pp.~1--20.

\bibitem{Nguyen2018Multi-taskRepresentation}
Nguyen, D.~K., and Okatani, T., 2018,
\newblock ``{Multi-task Learning of Hierarchical Vision-Language
  Representation},''
\newblock {\em Proceedings of the IEEE Computer Society Conference on Computer
  Vision and Pattern Recognition, {\bf 2019-June}}, 12, pp.~10484--10493.

\bibitem{Li2020Unicoder-VL:Pre-Training}
Li, G., Duan, N., Fang, Y., Gong, M., and Jiang, D., 2020,
\newblock ``{Unicoder-VL: A Universal Encoder for Vision and Language by
  Cross-Modal Pre-Training},''
\newblock {\em Proceedings of the AAAI Conference on Artificial Intelligence,
  {\bf 34}}(07), 4, pp.~11336--11344.

\bibitem{Su2019VL-BERT:Representations}
Su, W., Zhu, X., Cao, Y., Li, B., Lu, L., Wei, F., and Dai, J., 2019,
\newblock ``{VL-BERT: Pre-training of Generic Visual-Linguistic
  Representations},''.

\bibitem{Li2019VisualBERT:Language}
Li, L.~H., Yatskar, M., Yin, D., Hsieh, C.-J., and Chang, K.-W., 2019,
\newblock ``{VisualBERT: A Simple and Performant Baseline for Vision and
  Language},''.

\bibitem{Alberti2019FusionAnswering}
Alberti, C., Ling, J., Collins, M., and Reitter, D., 2019,
\newblock ``{Fusion of Detected Objects in Text for Visual Question
  Answering},''
\newblock {\em EMNLP-IJCNLP 2019 - 2019 Conference on Empirical Methods in
  Natural Language Processing and 9th International Joint Conference on Natural
  Language Processing, Proceedings of the Conference}, 8, pp.~2131--2140.

\bibitem{Sun2019VideoBERT:Learning}
Sun, C., Myers, A., Vondrick, C., Murphy, K., and Schmid, C., 2019,
\newblock ``{VideoBERT: A Joint Model for Video and Language Representation
  Learning},''
\newblock {\em Proceedings of the IEEE International Conference on Computer
  Vision}, 4, pp.~7463--7472.

\bibitem{Ngiam2011multimodalLearning}
Ngiam, J., Khosla, A., Kim, M., Nam, J., Lee, H., and Ng, A.~Y., 2011,
\newblock ``{Multimodal Deep Learning},''
\newblock In ICML 2011.

\bibitem{Silberer2014LearningAutoencoders}
Silberer, C., and Lapata, M., 2014,
\newblock ``{Learning Grounded Meaning Representations with Autoencoders},''
\newblock {\em 52nd Annual Meeting of the Association for Computational
  Linguistics, ACL 2014 - Proceedings of the Conference, {\bf 1}},
  pp.~721--732.

\bibitem{Feng2014Cross-modalAutoencoder}
Feng, F., Wang, X., and Li, R., 2014,
\newblock ``{Cross-modal retrieval with correspondence autoencoder},''
\newblock {\em MM 2014 - Proceedings of the 2014 ACM Conference on Multimedia},
  11, pp.~7--16.

\bibitem{Radford2021LearningSupervision}
Radford, A., Kim, J.~W., Hallacy, C., Ramesh, A., Goh, G., Agarwal, S., Sastry,
  G., Askell, A., Mishkin, P., Clark, J., Krueger, G., and Sutskever, I., 2021,
\newblock ``{Learning Transferable Visual Models From Natural Language
  Supervision},''.

\bibitem{Andrew2013DeepAnalysis}
Andrew, G., Arora, R., Bilmes, J., and Livescu, K., 2013,
\newblock {Deep Canonical Correlation Analysis}, 5.

\bibitem{Yang2017DeepData}
Yang, X., Ramesh, P., Chitta, R., Madhvanath, S., Bernal, E.~A., and Luo, J.,
  2017,
\newblock ``{Deep Multimodal Representation Learning from Temporal Data},''
\newblock In 2017 IEEE Conference on Computer Vision and Pattern Recognition
  (CVPR), pp.~5447--5455.

\bibitem{Bachman2019LearningViews}
Bachman, P., Hjelm, D., and Buchwalter, W., 2019,
\newblock ``{Learning Representations by Maximizing Mutual Information Across
  Views},''
\newblock In NIPS'19: Proceedings of the 33rd International Conference on
  Neural Information Processing Systems, pp.~15535--15545.

\bibitem{Zhang2020ContrastiveText}
Zhang, Y., Jiang, H., Miura, Y., Manning, C.~D., and Langlotz, C.~P., 2020,
\newblock ``{Contrastive Learning of Medical Visual Representations from Paired
  Images and Text},''
\newblock {\em Proceedings of Machine Learning Research, {\bf 182}}, 10,
  pp.~1--24.

\bibitem{Kiros2014UnifyingModels}
Kiros, R., Salakhutdinov, R., and Zemel, R.~S., 2014,
\newblock ``{Unifying Visual-Semantic Embeddings with Multimodal Neural
  Language Models},''
\newblock {\em undefined}.

\bibitem{Huang2013LearningData}
Huang, P.-S., He, X., Gao, J., Deng, L., Acero, A., and Heck, L., 2013,
\newblock {Learning Deep Structured Semantic Models for Web Search using
  Clickthrough Data}, 10.

\bibitem{Karpathy2014DeepDescriptions}
Karpathy, A., and Fei-Fei, L., 2014,
\newblock ``{Deep Visual-Semantic Alignments for Generating Image
  Descriptions},''
\newblock {\em IEEE Transactions on Pattern Analysis and Machine Intelligence,
  {\bf 39}}(4), 12, pp.~664--676.

\bibitem{Karpathy2014DeepMapping}
Karpathy, A., Joulin, A., and Fei-Fei, L., 2014,
\newblock ``{Deep Fragment Embeddings for Bidirectional Image Sentence
  Mapping},''
\newblock {\em Advances in Neural Information Processing Systems, {\bf
  3}}(January), 6, pp.~1889--1897.

\bibitem{Wu2019UnifiedRepresentations}
Wu, H., Mao, J., Zhang, Y., Jiang, Y., Li, L., Sun, W., and Ma, W.~Y., 2019,
\newblock ``{Unified visual-semantic embeddings: Bridging vision and language
  with structured meaning representations},''
\newblock {\em Proceedings of the IEEE Computer Society Conference on Computer
  Vision and Pattern Recognition, {\bf 2019-June}}, 6, pp.~6602--6611.

\bibitem{Plummer2015Flickr30kModels}
Plummer, B.~A., Wang, L., Cervantes, C.~M., Caicedo, J.~C., Hockenmaier, J.,
  and Lazebnik, S., 2015,
\newblock ``{Flickr30k Entities: Collecting Region-to-Phrase Correspondences
  for Richer Image-to-Sentence Models},''
\newblock {\em International Journal of Computer Vision, {\bf 123}}(1), 5,
  pp.~74--93.

\bibitem{Tan2019LXMERT:Transformers}
Tan, H., and Bansal, M., 2019,
\newblock ``{LXMERT: Learning Cross-Modality Encoder Representations from
  Transformers},''
\newblock {\em EMNLP-IJCNLP 2019 - 2019 Conference on Empirical Methods in
  Natural Language Processing and 9th International Joint Conference on Natural
  Language Processing, Proceedings of the Conference}, 8, pp.~5100--5111.

\bibitem{Lu2019ViLBERT:Tasks}
Lu, J., Batra, D., Parikh, D., and Lee, S., 2019,
\newblock ``{ViLBERT: Pretraining Task-Agnostic Visiolinguistic Representations
  for Vision-and-Language Tasks},''
\newblock {\em Advances in Neural Information Processing Systems, {\bf 32}}, 8.

\bibitem{Pramanik2019OmniNet:Learning}
Pramanik, S., Agrawal, P., and Hussain, A., 2019,
\newblock ``{OmniNet: A unified architecture for multi-modal multi-task
  learning},''.

\bibitem{Sbrolli2022IC3D:Generation}
Sbrolli, C., Cudrano, P., Frosi, M., and Matteucci, M., 2022,
\newblock ``{IC3D: Image-Conditioned 3D Diffusion for Shape Generation},''.

\bibitem{Nojavanasghari2016DeepPrediction}
Nojavanasghari, B., Gopinath, D., Koushik, J., Baltru{\v{s}}aitis, T., and
  Morency, L.~P., 2016,
\newblock ``{Deep multimodal fusion for persuasiveness prediction},''
\newblock {\em ICMI 2016 - Proceedings of the 18th ACM International Conference
  on Multimodal Interaction}, 10, pp.~284--288.

\bibitem{Anastasopoulos2019NeuralFeatures}
Anastasopoulos, A., Kumar, S., and Liao, H., 2019,
\newblock ``{Neural Language Modeling with Visual Features},''
\newblock {\em undefined}, 3.

\bibitem{Vielzeuf2019CentralNet:Fusion}
Vielzeuf, V., Lechervy, A., Pateux, S., and Jurie, F., 2019,
\newblock ``{CentralNet: A multilayer approach for multimodal fusion},''
\newblock {\em Lecture Notes in Computer Science (including subseries Lecture
  Notes in Artificial Intelligence and Lecture Notes in Bioinformatics), {\bf
  11134 LNCS}}, pp.~575--589.

\bibitem{10.1115/1.4054001}
Liu, B., Liu, X., Yang, Z., and Wang, C. C.~L., 2022,
\newblock ``{Concise and Effective Network for 3D Human Modeling From
  Orthogonal Silhouettes},''
\newblock {\em Journal of Computing and Information Science in Engineering,
  {\bf 22}}(5), 03, p.~051004.

\bibitem{Shutova2016BlackFeatures}
Shutova, E., Kiela, D., and Maillard, J., 2016,
\newblock ``{Black Holes and White Rabbits: Metaphor Identification with Visual
  Features},''
\newblock {\em 2016 Conference of the North American Chapter of the Association
  for Computational Linguistics: Human Language Technologies, NAACL HLT 2016 -
  Proceedings of the Conference}, pp.~160--170.

\bibitem{Cao2016DeepRetrieval}
Cao, Y., Long, M., Wang, J., Yang, Q., and Yuy, P.~S., 2016,
\newblock ``{Deep visual-semantic hashing for cross-modal retrieval},''
\newblock {\em Proceedings of the ACM SIGKDD International Conference on
  Knowledge Discovery and Data Mining, {\bf 13-17-Augu}}, 8, pp.~1445--1454.

\bibitem{Sikka2013MultipleWild}
Sikka, K., Dykstra, K., Sathyanarayana, S., Littlewort, G., and Bartlett, M.,
  2013,
\newblock ``{Multiple kernel learning for emotion recognition in the wild},''
\newblock {\em ICMI 2013 - Proceedings of the 2013 ACM International Conference
  on Multimodal Interaction}, pp.~517--524.

\bibitem{Morvant2014MajorityFusion}
Morvant, E., Habrard, A., and Ayache, S., 2014,
\newblock ``{Majority Vote of Diverse Classifiers for Late Fusion},''
\newblock {\em Lecture Notes in Computer Science (including subseries Lecture
  Notes in Artificial Intelligence and Lecture Notes in Bioinformatics), {\bf
  8621 LNCS}}, 4, pp.~153--162.

\bibitem{Perez-Rua2019MFAS:Search}
Perez-Rua, J.~M., Vielzeuf, V., Pateux, S., Baccouche, M., and Jurie, F., 2019,
\newblock ``{MFAS: Multimodal fusion architecture search},''
\newblock {\em Proceedings of the IEEE Computer Society Conference on Computer
  Vision and Pattern Recognition, {\bf 2019-June}}, 6, pp.~6959--6968.

\bibitem{Zhou2019EffectiveDiagnosis}
Zhou, T., Thung, K.~H., Zhu, X., and Shen, D., 2019,
\newblock ``{Effective feature learning and fusion of multimodality data using
  stage-wise deep neural network for dementia diagnosis},''
\newblock {\em Human brain mapping, {\bf 40}}(3), 2, pp.~1001--1016.

\bibitem{Zoph2016NeuralLearning}
Zoph, B., and Le, Q.~V., 2016,
\newblock ``{Neural Architecture Search with Reinforcement Learning},''
\newblock {\em 5th International Conference on Learning Representations, ICLR
  2017 - Conference Track Proceedings}, 11.

\bibitem{Tenenbaum2000SeparatingModels}
Tenenbaum, J.~B., and Freeman, W.~T., 2000,
\newblock ``{Separating style and content with bilinear models},''
\newblock {\em Neural Computation, {\bf 12}}(6), pp.~1247--1283.

\bibitem{Zadeh2017TensorAnalysis}
Zadeh, A., Chen, M., Cambria, E., Poria, S., and Morency, L.~P., 2017,
\newblock ``{Tensor Fusion Network for Multimodal Sentiment Analysis},''
\newblock {\em EMNLP 2017 - Conference on Empirical Methods in Natural Language
  Processing, Proceedings}, 7, pp.~1103--1114.

\bibitem{Chen2019PathomicPrognosis}
Chen, R.~J., Lu, M.~Y., Wang, J., Williamson, D.~F., Rodig, S.~J., Lindeman,
  N.~I., and Mahmood, F., 2019,
\newblock ``{Pathomic Fusion: An Integrated Framework for Fusing Histopathology
  and Genomic Features for Cancer Diagnosis and Prognosis},''
\newblock {\em IEEE Transactions on Medical Imaging, {\bf 41}}(4), 12,
  pp.~757--770.

\bibitem{Kim2022HadamardPooling}
Kim, J.-H., On, K.-W., Lim, W., Kim, J., Ha, J.-W., and Zhang, B.-T., 2022,
\newblock {Hadamard Product for Low-rank Bilinear Pooling}, 7.

\bibitem{Yu2017multi-modalAnswering}
Yu, Z., Yu, J., Fan, J., and Tao, D., 2017,
\newblock ``{Multi-modal Factorized Bilinear Pooling with Co-Attention Learning
  for Visual Question Answering},''
\newblock {\em Proceedings of the IEEE International Conference on Computer
  Vision, {\bf 2017-Octob}}, 8, pp.~1839--1848.

\bibitem{Yu2017BeyondAnswering}
Yu, Z., Yu, J., Xiang, C., Fan, J., and Tao, D., 2017,
\newblock ``{Beyond Bilinear: Generalized Multimodal Factorized High-order
  Pooling for Visual Question Answering},''
\newblock {\em IEEE Transactions on Neural Networks and Learning Systems, {\bf
  29}}(12), 8, pp.~5947--5959.

\bibitem{Gao2015CompactPooling}
Gao, Y., Beijbom, O., Zhang, N., and Darrell, T., 2015,
\newblock ``{Compact Bilinear Pooling},''
\newblock {\em Proceedings of the IEEE Computer Society Conference on Computer
  Vision and Pattern Recognition, {\bf 2016-Decem}}, 11, pp.~317--326.

\bibitem{Fukui2016multimodalGrounding}
Fukui, A., Park, D.~H., Yang, D., Rohrbach, A., Darrell, T., and Rohrbach, M.,
  2016,
\newblock ``{Multimodal Compact Bilinear Pooling for Visual Question Answering
  and Visual Grounding},''
\newblock {\em EMNLP 2016 - Conference on Empirical Methods in Natural Language
  Processing, Proceedings}, 6, pp.~457--468.

\bibitem{Ben-Younes2017MUTAN:Answering}
Ben-Younes, H., Cadene, R., Cord, M., and Thome, N., 2017,
\newblock ``{MUTAN: Multimodal Tucker Fusion for Visual Question Answering},''
\newblock {\em Proceedings of the IEEE International Conference on Computer
  Vision, {\bf 2017-Octob}}, 5, pp.~2631--2639.

\bibitem{Tucker1966SomeAnalysis}
Tucker, L.~R., 1966,
\newblock ``{Some mathematical notes on three-mode factor analysis},''
\newblock {\em Psychometrika 1966 31:3, {\bf 31}}(3), 9, pp.~279--311.

\bibitem{Ben-Younes2019BLOCK:Detection}
Ben-Younes, H., Cadene, R., Thome, N., and Cord, M., 2019,
\newblock ``{BLOCK: Bilinear Superdiagonal Fusion for Visual Question Answering
  and Visual Relationship Detection},''
\newblock {\em 33rd AAAI Conference on Artificial Intelligence, AAAI 2019, 31st
  Innovative Applications of Artificial Intelligence Conference, IAAI 2019 and
  the 9th AAAI Symposium on Educational Advances in Artificial Intelligence,
  EAAI 2019}, 1, pp.~8102--8109.

\bibitem{Jiang2022DeepClassification}
Jiang, S., Hu, J., Magee, C.~L., and Luo, J., 2022,
\newblock ``{Deep Learning for Technical Document Classification},''
\newblock {\em IEEE Transactions on Engineering Management}.

\bibitem{Parisot2018DiseaseDisease}
Parisot, S., Ktena, S.~I., Ferrante, E., Lee, M., Guerrero, R., Glocker, B.,
  and Rueckert, D., 2018,
\newblock ``{Disease prediction using graph convolutional networks: Application
  to Autism Spectrum Disorder and Alzheimer's disease},''
\newblock {\em Medical image analysis, {\bf 48}}, 8, pp.~117--130.

\bibitem{Cao2021UsingData}
Cao, M., Yang, M., Qin, C., Zhu, X., Chen, Y., Wang, J., and Liu, T., 2021,
\newblock ``{Using DeepGCN to identify the autism spectrum disorder from
  multi-site resting-state data},''
\newblock {\em Biomedical Signal Processing and Control, {\bf 70}}, 9,
  p.~103015.

\bibitem{Baltrusaitis2017multimodalTaxonomy}
Baltrusaitis, T., Ahuja, C., and Morency, L.~P., 2017,
\newblock ``{Multimodal Machine Learning: A Survey and Taxonomy},''
\newblock {\em IEEE Transactions on Pattern Analysis and Machine Intelligence,
  {\bf 41}}(2), 5, pp.~423--443.

\bibitem{Vaswani2017AttentionNeed}
Vaswani, A., Shazeer, N., Parmar, N., Uszkoreit, J., Jones, L., Gomez, A.~N.,
  Kaiser, L., and Polosukhin, I., 2017,
\newblock ``{Attention is all you need},''
\newblock In Advances in Neural Information Processing Systems,
  Vol.~2017-Decem, Neural information processing systems foundation,
  pp.~5999--6009.

\bibitem{Graves2014NeuralMachines}
Graves, A., Wayne, G., and Danihelka, I., 2014,
\newblock ``{Neural Turing Machines},''
\newblock {\em arXiv preprint arXiv:1410.5401.}, 10.

\bibitem{Bahdanau2014NeuralTranslate}
Bahdanau, D., Cho, K., and Bengio, Y., 2014,
\newblock ``{Neural Machine Translation by Jointly Learning to Align and
  Translate},''
\newblock {\em 3rd International Conference on Learning Representations, ICLR
  2015 - Conference Track Proceedings}, 9.

\bibitem{Zhu2016Visual7W:Images}
Zhu, Y., Groth, O., Bernstein, M., and Fei-Fei, L., 2016,
\newblock ``{Visual7W: Grounded question answering in images},''
\newblock {\em Proceedings of the IEEE Computer Society Conference on Computer
  Vision and Pattern Recognition, {\bf 2016-Decem}}, 12, pp.~4995--5004.

\bibitem{Shih2015WhereAnswering}
Shih, K.~J., Singh, S., and Hoiem, D., 2015,
\newblock ``{Where To Look: Focus Regions for Visual Question Answering},''
\newblock {\em Proceedings of the IEEE Computer Society Conference on Computer
  Vision and Pattern Recognition, {\bf 2016-Decem}}, 11, pp.~4613--4621.

\bibitem{Xu2015AskAnswering}
Xu, H., and Saenko, K., 2015,
\newblock ``{Ask, Attend and Answer: Exploring Question-Guided Spatial
  Attention for Visual Question Answering},''
\newblock {\em Lecture Notes in Computer Science (including subseries Lecture
  Notes in Artificial Intelligence and Lecture Notes in Bioinformatics), {\bf
  9911 LNCS}}, 11, pp.~451--466.

\bibitem{Anderson2017Bottom-UpAnswering}
Anderson, P., He, X., Buehler, C., Teney, D., Johnson, M., Gould, S., and
  Zhang, L., 2017,
\newblock ``{Bottom-Up and Top-Down Attention for Image Captioning and Visual
  Question Answering},''
\newblock {\em Proceedings of the IEEE Computer Society Conference on Computer
  Vision and Pattern Recognition}, 7, pp.~6077--6086.

\bibitem{Mansimov2015GeneratingAttention}
Mansimov, E., Parisotto, E., Ba, J.~L., and Salakhutdinov, R., 2015,
\newblock ``{Generating Images from Captions with Attention},''
\newblock {\em 4th International Conference on Learning Representations, ICLR
  2016 - Conference Track Proceedings}, 11.

\bibitem{Xu2018AttnGAN:Networks}
Xu, T., Zhang, P., Huang, Q., Zhang, H., Gan, Z., Huang, X., and He, X., 2018,
\newblock ``{AttnGAN: Fine-Grained Text to Image Generation With Attentional
  Generative Adversarial Networks},''
\newblock In Proceedings of the IEEE Conference on Computer Vision and Pattern
  Recognition (CVPR), pp.~1316--1324.

\bibitem{Li2019Object-drivenTraining}
Li, W., Zhang, P., Zhang, L., Huang, Q., He, X., Lyu, S., and Gao, J., 2019,
\newblock ``{Object-driven Text-to-Image Synthesis via Adversarial Training},''
\newblock {\em Proceedings of the IEEE Computer Society Conference on Computer
  Vision and Pattern Recognition, {\bf 2019-June}}, 2, pp.~12166--12174.

\bibitem{Nam2017DualMatching}
Nam, H., Ha, J.-W., and Kim, J., 2017,
\newblock ``{Dual Attention Networks for Multimodal Reasoning and Matching},''
\newblock {\em 2017 IEEE Conference on Computer Vision and Pattern Recognition
  (CVPR)}, 7, pp.~2156--2164.

\bibitem{Lu2016HierarchicalAnswering}
Lu, J., Yang, J., Batra, D., and Parikh, D., 2016,
\newblock ``{Hierarchical Question-Image Co-Attention for Visual Question
  Answering},''
\newblock In NIPS'16: Proceedings of the 30th International Conference on
  Neural Information Processing Systems, pp.~289--297.

\bibitem{Osman2018DualAnswering}
Osman, A., and Samek, W., 2018,
\newblock ``{Dual Recurrent Attention Units for Visual Question Answering},''
\newblock {\em Computer Vision and Image Understanding, {\bf 185}}, 2,
  pp.~24--30.

\bibitem{Schwartz2017High-OrderAnswering}
Schwartz, I., Schwing, A.~G., and Hazan, T., 2017,
\newblock ``{High-Order Attention Models for Visual Question Answering},''
\newblock {\em Advances in Neural Information Processing Systems, {\bf
  2017-Decem}}, 11, pp.~3665--3675.

\bibitem{Yang2015StackedAnswering}
Yang, Z., He, X., Gao, J., Deng, L., and Smola, A., 2015,
\newblock ``{Stacked Attention Networks for Image Question Answering},''
\newblock {\em Proceedings of the IEEE Computer Society Conference on Computer
  Vision and Pattern Recognition, {\bf 2016-Decem}}, 11, pp.~21--29.

\bibitem{Fan2018StackedReasoning}
Fan, H., and Zhou, J., 2018,
\newblock ``{Stacked Latent Attention for Multimodal Reasoning},''
\newblock {\em Proceedings of the IEEE Computer Society Conference on Computer
  Vision and Pattern Recognition}, 12, pp.~1072--1080.

\bibitem{Xiong2016DynamicAnswering}
Xiong, C., Merity, S., and Socher, R., 2016,
\newblock ``{Dynamic Memory Networks for Visual and Textual Question
  Answering},''
\newblock {\em 33rd International Conference on Machine Learning, ICML 2016,
  {\bf 5}}, 3, pp.~3574--3583.

\bibitem{Ren2015FasterNetworks}
Ren, S., He, K., Girshick, R., and Sun, J., 2015,
\newblock ``{Faster R-CNN: Towards Real-Time Object Detection with Region
  Proposal Networks},''
\newblock {\em Advances in Neural Information Processing Systems, {\bf 28}}, 6.

\bibitem{Lu2017Co-attendingAnswering}
Lu, P., Li, H., Zhang, W., Wang, J., and Wang, X., 2017,
\newblock ``{Co-attending Free-form Regions and Detections with Multi-modal
  Multiplicative Feature Embedding for Visual Question Answering},''
\newblock {\em 32nd AAAI Conference on Artificial Intelligence, AAAI 2018}, 11,
  pp.~7218--7225.

\bibitem{Rombach2021High-ResolutionModels}
Rombach, R., Blattmann, A., Lorenz, D., Esser, P., and Ommer, B., 2021,
\newblock ``{High-Resolution Image Synthesis with Latent Diffusion Models},''
\newblock pp.~10674--10685.

\bibitem{data2vec}
Baevski, A., Hsu, W.-N., Xu, Q., Babu, A., Gu, J., and Auli, M., 2022,
\newblock data2vec: A general framework for self-supervised learning in speech,
  vision and language.

\bibitem{Kim2016multimodalQA}
Kim, J.~H., Lee, S.~W., Kwak, D., Heo, M.~O., Kim, J., Ha, J.~W., and Zhang,
  B.~T., 2016,
\newblock ``{Multimodal Residual Learning for Visual QA},''
\newblock {\em Advances in Neural Information Processing Systems}, 6,
  pp.~361--369.

\bibitem{Arevalo2017GatedFusion}
Arevalo, J., Solorio, T., Montes-Y-G{\'{o}}mez, M., and Gonz{\'{a}}lez, F.~A.,
  2017,
\newblock ``{Gated Multimodal Units for Information Fusion},''
\newblock {\em 5th International Conference on Learning Representations, ICLR
  2017 - Workshop Track Proceedings}, 2.

\bibitem{Noh2015ImagePrediction}
Noh, H., Seo, P.~H., and Han, B., 2015,
\newblock ``{Image Question Answering using Convolutional Neural Network with
  Dynamic Parameter Prediction},''
\newblock {\em Proceedings of the IEEE Computer Society Conference on Computer
  Vision and Pattern Recognition, {\bf 2016-Decem}}, 11, pp.~30--38.

\bibitem{Tolstikhin2014GenerativeNetworks}
Tolstikhin, I., Bousquet, O., Sch{\"{o}}lkopf, B., Thierbach, K., Bazin, P.~L.,
  de~Back, W., Gavriilidis, F., Kirilina, E., J{\"{a}}ger, C., Morawski, M.,
  Geyer, S., Weiskopf, N., Scherf, N., Goodfellow, I.~J., Pouget-Abadie, J.,
  Mirza, M., Xu, B., Warde-Farley, D., Ozair, S., Courville, A., Bengio, Y.,
  Musk, E., {Neuralink}, Hjorts{\o}, M.~A., Wolenski, P., Ruder, S., Grathwohl,
  W., Chen, R. T.~Q., Bettencourt, J., Sutskever, I., Duvenaud, D., and
  Doersch, C., 2014,
\newblock ``{Generative Adversarial Networks},''
\newblock {\em Lecture Notes in Computer Science (including subseries Lecture
  Notes in Artificial Intelligence and Lecture Notes in Bioinformatics), {\bf
  11046 LNCS}}(NeurIPS), 6, pp.~1--9.

\bibitem{Mirza2014ConditionalNets}
Mirza, M., and Osindero, S., 2014,
\newblock ``{Conditional Generative Adversarial Nets},''.

\bibitem{Reed2016GenerativeSynthesis}
Reed, S., Akata, Z., Yan, X., Logeswaran, L., Schiele, B., and Lee, H., 2016,
\newblock ``{Generative Adversarial Text to Image Synthesis},''
\newblock {\em 33rd International Conference on Machine Learning, ICML 2016,
  {\bf 3}}, 5, pp.~1681--1690.

\bibitem{Zhang2016StackGAN:Networks}
Zhang, H., Xu, T., Li, H., Zhang, S., Wang, X., Huang, X., and Metaxas, D.,
  2016,
\newblock ``{StackGAN: Text to Photo-realistic Image Synthesis with Stacked
  Generative Adversarial Networks},''
\newblock pp.~5908--5916.

\bibitem{Zhang2019StackGAN++:Networks}
Zhang, H., Xu, T., Li, H., Zhang, S., Wang, X., Huang, X., and Metaxas, D.~N.,
  2019,
\newblock ``{StackGAN++: Realistic Image Synthesis with Stacked Generative
  Adversarial Networks},''
\newblock {\em IEEE Transactions on Pattern Analysis and Machine Intelligence,
  {\bf 41}}(08), 8, pp.~1947--1962.

\bibitem{Zhu2019DM-GAN:Synthesis}
Zhu, M., Pan, P., Chen, W., and Yang, Y., 2019,
\newblock ``{DM-GAN: Dynamic Memory Generative Adversarial Networks for
  Text-to-Image Synthesis},''
\newblock {\em Proceedings of the IEEE Computer Society Conference on Computer
  Vision and Pattern Recognition, {\bf 2019-June}}, 4, pp.~5795--5803.

\bibitem{Zhang2018PhotographicNetwork}
Zhang, Z., Xie, Y., and Yang, L., 2018,
\newblock ``{Photographic Text-to-Image Synthesis with a Hierarchically-nested
  Adversarial Network},''
\newblock {\em Proceedings of the IEEE Computer Society Conference on Computer
  Vision and Pattern Recognition}, 2, pp.~6199--6208.

\bibitem{Dash2017TAC-GANNetworkb}
Dash, A., Gamboa, J. C.~B., Ahmed, S., Liwicki, M., and Afzal, M.~Z., 2017,
\newblock ``{TAC-GAN - Text Conditioned Auxiliary Classifier Generative
  Adversarial Network},''
\newblock In Proc. CVPR.

\bibitem{Cha2018AdversarialSynthesis}
Cha, M., Gwon, Y.~L., and Kung, H.~T., 2018,
\newblock ``{Adversarial Learning of Semantic Relevance in Text to Image
  Synthesis},''
\newblock {\em 33rd AAAI Conference on Artificial Intelligence, AAAI 2019, 31st
  Innovative Applications of Artificial Intelligence Conference, IAAI 2019 and
  the 9th AAAI Symposium on Educational Advances in Artificial Intelligence,
  EAAI 2019}, 12, pp.~3272--3279.

\bibitem{Qiao2019MirrorGAN:Redescription}
Qiao, T., Zhang, J., Xu, D., and Tao, D., 2019,
\newblock ``{MirrorGAN: Learning Text-to-image Generation by Redescription},''
\newblock {\em Proceedings of the IEEE Computer Society Conference on Computer
  Vision and Pattern Recognition, {\bf 2019-June}}, 3, pp.~1505--1514.

\bibitem{Reed2016LearningDraw}
Reed, S., Akata, Z., Mohan, S., Tenka, S., Schiele, B., and Lee, H., 2016,
\newblock ``{Learning What and Where to Draw},''
\newblock {\em Advances in Neural Information Processing Systems}, 10,
  pp.~217--225.

\bibitem{Zhao2018ImageLayout}
Zhao, B., Meng, L., Yin, W., and Sigal, L., 2018,
\newblock ``{Image Generation from Layout},''
\newblock {\em Proceedings of the IEEE Computer Society Conference on Computer
  Vision and Pattern Recognition, {\bf 2019-June}}, 11, pp.~8576--8585.

\bibitem{Hinz2019GeneratingLocations}
Hinz, T., Heinrich, S., and Wermter, S., 2019,
\newblock ``{Generating Multiple Objects at Spatially Distinct Locations},''
\newblock {\em 7th International Conference on Learning Representations, ICLR
  2019}, 1.

\bibitem{Hong2018InferringSynthesis}
Hong, S., Yang, D., Choi, J., and Lee, H., 2018,
\newblock ``{Inferring Semantic Layout for Hierarchical Text-to-Image
  Synthesis},''
\newblock {\em Proceedings of the IEEE Computer Society Conference on Computer
  Vision and Pattern Recognition}, 1, pp.~7986--7994.

\bibitem{Johnson2018ImageGraphs}
Johnson, J., Gupta, A., and Fei-Fei, L., 2018,
\newblock ``{Image Generation from Scene Graphs},''
\newblock {\em Proceedings of the IEEE Computer Society Conference on Computer
  Vision and Pattern Recognition}, 4, pp.~1219--1228.

\bibitem{Mao2014Deepm-RNN}
Mao, J., Xu, W., Yang, Y., Wang, J., Huang, Z., and Yuille, A., 2014,
\newblock ``{Deep Captioning with Multimodal Recurrent Neural Networks
  (m-RNN)},''
\newblock {\em 3rd International Conference on Learning Representations, ICLR
  2015 - Conference Track Proceedings}, 12.

\bibitem{vandenOordDeepMind2017NeuralLearning}
van~den Oord~DeepMind, A., Vinyals~DeepMind, O., and Kavukcuoglu~DeepMind, K.,
  2017,
\newblock ``{Neural Discrete Representation Learning},''
\newblock {\em Advances in Neural Information Processing Systems, {\bf 30}}.

\bibitem{Sanghi2021CLIP-Forge:Generation}
Sanghi, A., Chu, H., Lambourne, J.~G., Wang, Y., Cheng, C.-Y., Fumero, M., and
  Malekshan, K.~R., 2021,
\newblock ``{CLIP-Forge: Towards Zero-Shot Text-to-Shape Generation},''.

\bibitem{Shetty2017SpeakingTraining}
Shetty, R., Rohrbach, M., Hendricks, L.~A., Fritz, M., and Schiele, B., 2017,
\newblock ``{Speaking the Same Language: Matching Machine to Human Captions by
  Adversarial Training},''
\newblock {\em Proceedings of the IEEE International Conference on Computer
  Vision, {\bf 2017-Octob}}, 3, pp.~4155--4164.

\bibitem{Ajit2020ANetworks}
Ajit, A., Acharya, K., and Samanta, A., 2020,
\newblock ``{A Review of Convolutional Neural Networks},''
\newblock {\em International Conference on Emerging Trends in Information
  Technology and Engineering, ic-ETITE 2020}, 2.

\bibitem{Li2021AProspects}
Li, Z., Liu, F., Yang, W., Peng, S., and Zhou, J., 2021,
\newblock ``{A Survey of Convolutional Neural Networks: Analysis, Applications,
  and Prospects},''
\newblock {\em IEEE Transactions on Neural Networks and Learning Systems}, 6,
  pp.~1--21.

\bibitem{Fathi2018DeepProcessing}
Fathi, E., and Maleki~Shoja, B., 2018,
\newblock ``{Deep Neural Networks for Natural Language Processing},''
\newblock {\em Handbook of Statistics, {\bf 38}}, 1, pp.~229--316.

\bibitem{Mikolov2013DistributedCompositionality}
Mikolov, T., Chen, K., Corrado, G.~S., Dean, J., Sutskever, I., Chen, K.,
  Corrado, G.~S., and Dean, J., 2013,
\newblock ``{Distributed Representations of Words and Phrases and their
  Compositionality},''
\newblock {\em Advances in Neural Information Processing Systems}, 10,
  pp.~1--9.

\bibitem{Yagcioglu2015ACaptioning}
Yagcioglu, S., Erdem, E., Erdem, A., and {\c{C}}akici, R., 2015,
\newblock ``{A Distributed Representation Based Query Expansion Approach for
  Image Captioning},''
\newblock {\em ACL-IJCNLP 2015 - 53rd Annual Meeting of the Association for
  Computational Linguistics and the 7th International Joint Conference on
  Natural Language Processing of the Asian Federation of Natural Language
  Processing, Proceedings of the Conference, {\bf 2}}, pp.~106--111.

\bibitem{Cordonnier2019OnLayers}
Cordonnier, J.-B., Loukas, A., and Jaggi, M., 2019,
\newblock ``{On the Relationship between Self-Attention and Convolutional
  Layers},''.

\bibitem{Dosovitskiy2020AnScale}
Dosovitskiy, A., Beyer, L., Kolesnikov, A., Weissenborn, D., Zhai, X.,
  Unterthiner, T., Dehghani, M., Minderer, M., Heigold, G., Gelly, S.,
  Uszkoreit, J., and Houlsby, N., 2020,
\newblock ``{An Image is Worth 16x16 Words: Transformers for Image Recognition
  at Scale},''.

\bibitem{Wang2022End-to-EndCaptioning}
Wang, Y., Xu, J., and Sun, Y., 2022,
\newblock ``{End-to-End Transformer Based Model for Image Captioning},''
\newblock {\em Proceedings of the AAAI Conference on Artificial Intelligence,
  {\bf 36}}(3), 3, pp.~2585--2594.

\bibitem{Kalyan2021AMMUSProcessing}
Kalyan, K.~S., Rajasekharan, A., and Sangeetha, S., 2021,
\newblock ``{AMMUS : A Survey of Transformer-based Pretrained Models in Natural
  Language Processing},''.

\bibitem{Sohl-Dickstein2015DeepThermodynamics}
Sohl-Dickstein, J., Weiss, E.~A., Maheswaranathan, N., and Ganguli, S., 2015,
\newblock ``{Deep Unsupervised Learning using Nonequilibrium Thermodynamics},''
\newblock {\em 32nd International Conference on Machine Learning, ICML 2015,
  {\bf 3}}, 3, pp.~2246--2255.

\bibitem{Ho2020DenoisingModels}
Ho, J., Jain, A., and Abbeel, P., 2020,
\newblock ``{Denoising Diffusion Probabilistic Models},''
\newblock {\em Advances in Neural Information Processing Systems, {\bf
  2020-Decem}}, 6.

\bibitem{Song2020DenoisingModels}
Song, J., Meng, C., and Ermon, S., 2020,
\newblock ``{Denoising Diffusion Implicit Models},''.

\bibitem{Song2020Score-BasedEquations}
Song, Y., Sohl-Dickstein, J., Kingma, D.~P., Kumar, A., Ermon, S., and Poole,
  B., 2020,
\newblock ``{Score-Based Generative Modeling through Stochastic Differential
  Equations},''.

\bibitem{Vahdat2021Score-basedSpace}
Vahdat, A., Kreis, K., and Kautz, J., 2021,
\newblock ``{Score-based Generative Modeling in Latent Space},''
\newblock {\em Advances in Neural Information Processing Systems, {\bf 14}}, 6,
  pp.~11287--11302.

\bibitem{Luo2021DiffusionGeneration}
Luo, S., and Hu, W., 2021,
\newblock ``{Diffusion Probabilistic Models for 3D Point Cloud Generation},''
\newblock {\em Proceedings of the IEEE Computer Society Conference on Computer
  Vision and Pattern Recognition}, 3, pp.~2836--2844.

\bibitem{Zhou20213DDiffusion}
Zhou, L., Du, Y., and Wu, J., 2021,
\newblock ``{3D Shape Generation and Completion through Point-Voxel
  Diffusion},''
\newblock {\em Proceedings of the IEEE International Conference on Computer
  Vision}, 4, pp.~5806--5815.

\bibitem{Zeng2022LION:Generation}
Zeng, X., Vahdat, A., Williams, F., Gojcic, Z., Litany, O., Fidler, S., and
  Kreis, K., 2022,
\newblock ``{LION: Latent Point Diffusion Models for 3D Shape Generation},''.

\bibitem{Liu2019Point-VoxelLearning}
Liu, Z., Tang, H., Lin, Y., and Han, S., 2019,
\newblock ``{Point-Voxel CNN for Efficient 3D Deep Learning},''
\newblock {\em Advances in Neural Information Processing Systems, {\bf 32}}, 7.

\bibitem{Ho2022Classifier-FreeGuidance}
Ho, J., and Salimans, T., 2022,
\newblock ``{Classifier-Free Diffusion Guidance},''.

\bibitem{Nichol2022Point-E:Prompts}
Nichol, A., Jun, H., Dhariwal, P., Mishkin, P., and Chen, M., 2022,
\newblock ``{Point-E: A System for Generating 3D Point Clouds from Complex
  Prompts},''.

\bibitem{Ramesh2022HierarchicalLatents}
Ramesh, A., Dhariwal, P., Nichol, A., Chu, C., and Chen, M., 2022,
\newblock ``{Hierarchical Text-Conditional Image Generation with CLIP
  Latents},''.

\bibitem{Mao2015GenerationDescriptions}
Mao, J., Huang, J., Toshev, A., Camburu, O., Yuille, A., and Murphy, K., 2015,
\newblock ``{Generation and Comprehension of Unambiguous Object
  Descriptions},''
\newblock {\em Proceedings of the IEEE Computer Society Conference on Computer
  Vision and Pattern Recognition, {\bf 2016-Decem}}, 11, pp.~11--20.

\bibitem{Vinyals2014ShowGenerator}
Vinyals, O., Toshev, A., Bengio, S., and Erhan, D., 2014,
\newblock ``{Show and Tell: A Neural Image Caption Generator},''
\newblock {\em Proceedings of the IEEE Computer Society Conference on Computer
  Vision and Pattern Recognition, {\bf 07-12-June}}, 11, pp.~3156--3164.

\bibitem{Rohrbach2015TheDescription}
Rohrbach, A., Rohrbach, M., and Schiele, B., 2015,
\newblock ``{The Long-Short Story of Movie Description},''
\newblock {\em Lecture Notes in Computer Science (including subseries Lecture
  Notes in Artificial Intelligence and Lecture Notes in Bioinformatics), {\bf
  9358}}, 6, pp.~209--221.

\bibitem{Yu2022ScalingGeneration}
Yu, J., Xu, Y., Koh, J.~Y., Luong, T., Baid, G., Wang, Z., Vasudevan, V., Ku,
  A., Yang, Y., Ayan, B.~K., Hutchinson, B., Han, W., Parekh, Z., Li, X.,
  Zhang, H., Baldridge, J., and Wu, Y., 2022,
\newblock ``{Scaling Autoregressive Models for Content-Rich Text-to-Image
  Generation},''.

\bibitem{Ding2021CogView:Transformers}
Ding, M., Yang, Z., Hong, W., Zheng, W., Zhou, C., Yin, D., Lin, J., Zou, X.,
  Shao, Z., Yang, H., and Tang, J., 2021,
\newblock ``{CogView: Mastering Text-to-Image Generation via Transformers},''
\newblock {\em Advances in Neural Information Processing Systems, {\bf 24}}, 5,
  pp.~19822--19835.

\bibitem{Desai2020VirTex:Annotations}
Desai, K., and Johnson, J., 2020,
\newblock ``{VirTex: Learning Visual Representations from Textual
  Annotations},''
\newblock {\em Proceedings of the IEEE Computer Society Conference on Computer
  Vision and Pattern Recognition}, 6, pp.~11157--11168.

\bibitem{BulentSariyildiz2020LearningAnnotations}
Bulent~Sariyildiz, M., Perez, J., Larlus, D., Sariyildiz, M.~B., Perez, J., and
  Larlus, D., 2020,
\newblock ``{Learning Visual Representations with Caption Annotations},''
\newblock {\em Lecture Notes in Computer Science (including subseries Lecture
  Notes in Artificial Intelligence and Lecture Notes in Bioinformatics), {\bf
  12353 LNCS}}, 8, pp.~153--170.

\bibitem{Dinh2016DensityNVP}
Dinh, L., Sohl-Dickstein~Google, J., Samy, B., and Google~Brain, B., 2016,
\newblock {Density estimation using Real NVP}, 7.

\bibitem{Wei2022Flow-basedImage}
Wei, Y., Vosselman, G., and Yang, M.~Y., 2022,
\newblock ``Flow-based gan for 3d point cloud generation from a single
  image,''.

\bibitem{Chen2018LearningModeling}
Chen, Z., and Zhang, H., 2018,
\newblock ``{Learning Implicit Fields for Generative Shape Modeling},''
\newblock {\em Proceedings of the IEEE Computer Society Conference on Computer
  Vision and Pattern Recognition, {\bf 2019-June}}, 12, pp.~5932--5941.

\bibitem{Liu2019LearningSupervision}
Liu, S., Saito, S., Chen, W., and Li, H., 2019,
\newblock ``{Learning to Infer Implicit Surfaces without 3D Supervision},''
\newblock {\em Advances in Neural Information Processing Systems, {\bf 32}},
  11.

\bibitem{Park2019DeepSDF:Representation}
Park, J.~J., Florence, P., Straub, J., Newcombe, R., and Lovegrove, S., 2019,
\newblock ``{DeepSDF: Learning Continuous Signed Distance Functions for Shape
  Representation},''
\newblock {\em Proceedings of the IEEE Computer Society Conference on Computer
  Vision and Pattern Recognition, {\bf 2019-June}}, 1, pp.~165--174.

\bibitem{Salimans2016ImprovedGANs}
Salimans, T., Goodfellow, I., Zaremba, W., Cheung, V., Radford, A., and Chen,
  X., 2016,
\newblock ``{Improved Techniques for Training GANs},''
\newblock {\em Advances in Neural Information Processing Systems}, 6,
  pp.~2234--2242.

\bibitem{Heusel2017GANsEquilibrium}
Heusel, M., Ramsauer, H., Unterthiner, T., Nessler, B., and Hochreiter, S.,
  2017,
\newblock ``{GANs Trained by a Two Time-Scale Update Rule Converge to a Local
  Nash Equilibrium},''
\newblock {\em Advances in Neural Information Processing Systems, {\bf
  2017-Decem}}, 6, pp.~6627--6638.

\bibitem{Odena2016ConditionalGANs}
Odena, A., Olah, C., and Shlens, J., 2016,
\newblock ``{Conditional Image Synthesis With Auxiliary Classifier GANs},''
\newblock {\em 34th International Conference on Machine Learning, ICML 2017,
  {\bf 6}}, 10, pp.~4043--4055.

\bibitem{Li2019ManiGAN:Manipulation}
Li, B., Qi, X., Lukasiewicz, T., and Torr, P.~H., 2019,
\newblock ``{ManiGAN: Text-Guided Image Manipulation},''
\newblock {\em Proceedings of the IEEE Computer Society Conference on Computer
  Vision and Pattern Recognition}, 12, pp.~7877--7886.

\bibitem{Achlioptas2017LearningClouds}
Achlioptas, P., Diamanti, O., Mitliagkas, I., and Guibas, L., 2017,
\newblock ``{Learning Representations and Generative Models for 3D Point
  Clouds},''
\newblock {\em 35th International Conference on Machine Learning, ICML 2018,
  {\bf 1}}, 7, pp.~67--85.

\bibitem{Shu20193DConvolutions}
Shu, D., Park, S.~W., and Kwon, J., 2019,
\newblock ``{3D Point Cloud Generative Adversarial Network Based on Tree
  Structured Graph Convolutions},''
\newblock {\em Proceedings of the IEEE International Conference on Computer
  Vision, {\bf 2019-Octob}}, 5, pp.~3858--3867.

\bibitem{Ibing20213DFunctions}
Ibing, M., Lim, I., and Kobbelt, L., 2021,
\newblock ``{3D Shape Generation with Grid-based Implicit Functions},''
\newblock {\em Proceedings of the IEEE Computer Society Conference on Computer
  Vision and Pattern Recognition}, 7, pp.~13554--13563.

\bibitem{Socher2013Zero-ShotTransfer}
Socher, R., Ganjoo, M., Sridhar, H., Bastani, O., Manning, C.~D., and Ng,
  A.~Y., 2013,
\newblock ``{Zero-Shot Learning Through Cross-Modal Transfer},''
\newblock {\em 1st International Conference on Learning Representations, ICLR
  2013 - Workshop Track Proceedings}, 1.

\bibitem{Tsai2018LearningRepresentations}
Tsai, Y.-H.~H., Liang, P.~P., Zadeh, A., Morency, L.-P., and Salakhutdinov, R.,
  2018,
\newblock ``{Learning Factorized Multimodal Representations},''
\newblock {\em 7th International Conference on Learning Representations, ICLR
  2019}, 6.

\bibitem{BaPredictingDescriptions}
Ba, J.~L., Swersky, K., Fidler, S., and Salakhutdinov, R.
\newblock ``{Predicting Deep Zero-Shot Convolutional Neural Networks using
  Textual Descriptions},''.

\bibitem{Reed2016LearningDescriptions}
Reed, S., Akata, Z., Lee, H., and Schiele, B., 2016,
\newblock ``{Learning Deep Representations of Fine-grained Visual
  Descriptions},''
\newblock {\em Proceedings of the IEEE Computer Society Conference on Computer
  Vision and Pattern Recognition, {\bf 2016-Decem}}, 5, pp.~49--58.

\bibitem{Nakov2009ImprovedLanguages}
Nakov, P., and Ng, H.~T., 2009,
\newblock {Improved Statistical Machine Translation for Resource-Poor Languages
  Using Related Resource-Rich Languages}.

\bibitem{Hendricks2015DeepData}
Hendricks, L.~A., Venugopalan, S., Rohrbach, M., Mooney, R., Saenko, K., and
  Darrell, T., 2015,
\newblock ``{Deep Compositional Captioning: Describing Novel Object Categories
  without Paired Training Data},''
\newblock {\em Proceedings of the IEEE Computer Society Conference on Computer
  Vision and Pattern Recognition, {\bf 2016-Decem}}, 11, pp.~1--10.

\bibitem{Socher2010ConnectingCorpora}
Socher, R., and Fei-Fei, L., 2010,
\newblock ``{Connecting modalities: Semi-supervised segmentation and annotation
  of images using unaligned text corpora},''
\newblock {\em Proceedings of the IEEE Computer Society Conference on Computer
  Vision and Pattern Recognition}, pp.~966--973.

\bibitem{Socher2014GroundedSentences}
Socher, R., Karpathy, A., Le, Q.~V., Manning, C.~D., and Ng, A.~Y., 2014,
\newblock ``{Grounded Compositional Semantics for Finding and Describing Images
  with Sentences},''
\newblock {\em Transactions of the Association for Computational Linguistics,
  {\bf 2}}, 12, pp.~207--218.

\bibitem{Feng2010VisualRepresentation}
Feng, Y., and Lapata, M., 2010,
\newblock {Visual Information in Semantic Representation}.

\bibitem{Bruni2012DistributionalTechnicolor}
Bruni, E., Boleda, G., Baroni, M., and Tran, N.-K., 2012,
\newblock {Distributional Semantics in Technicolor}.

\bibitem{Kottur2016VisualWord2VecScenes}
Kottur, S., Vedantam, R., Moura, J. M.~F., and Parikh, D., 2016,
\newblock ``{VisualWord2Vec (Vis-W2V): Learning Visually Grounded Word
  Embeddings Using Abstract Scenes},''
\newblock {\em 2016 IEEE Conference on Computer Vision and Pattern Recognition
  (CVPR)}, 6, pp.~4985--4994.

\bibitem{Gupta2019ViCo:Co-occurrences}
Gupta, T., Schwing, A., and Hoiem, D., 2019,
\newblock ``{ViCo: Word Embeddings from Visual Co-occurrences},''
\newblock {\em Proceedings of the IEEE International Conference on Computer
  Vision, {\bf 2019-Octob}}, 8, pp.~7424--7433.

\bibitem{Mori1999Image-to-wordDividing}
Mori, Y., Takahashi, H., and Oka, R., 1999,
\newblock ``{Image-to-word transformation based on dividing},''.

\bibitem{Quattoni2007LearningCaptions}
Quattoni, A., Collins, M., and Darrell, T., 2007,
\newblock ``{Learning visual representations using images with captions},''
\newblock {\em Proceedings of the IEEE Computer Society Conference on Computer
  Vision and Pattern Recognition}.

\bibitem{Joulin2015LearningData}
Joulin, A., van Der~Maaten, L., Jabri, A., and Vasilache, N., 2015,
\newblock ``{Learning Visual Features from Large Weakly Supervised Data},''
\newblock {\em Lecture Notes in Computer Science (including subseries Lecture
  Notes in Artificial Intelligence and Lecture Notes in Bioinformatics), {\bf
  9911 LNCS}}, 11, pp.~67--84.

\bibitem{Li2016LearningData}
Li, A., Jabri, A., Joulin, A., and Maaten, L. V.~D., 2016,
\newblock ``{Learning Visual N-Grams from Web Data},''
\newblock {\em Proceedings of the IEEE International Conference on Computer
  Vision, {\bf 2017-Octob}}, 12, pp.~4193--4202.

\bibitem{Mahajan2018ExploringPretraining}
Mahajan, D., Girshick, R., Ramanathan, V., He, K., Paluri, M., Li, Y.,
  Bharambe, A., and van~der Maaten, L., 2018,
\newblock ``{Exploring the Limits of Weakly Supervised Pretraining},''
\newblock {\em Lecture Notes in Computer Science (including subseries Lecture
  Notes in Artificial Intelligence and Lecture Notes in Bioinformatics), {\bf
  11206 LNCS}}, 5, pp.~185--201.

\bibitem{Kiela2015GroundingPerception}
Kiela, D., Bulat, L., and Clark, S., 2015,
\newblock ``{Grounding Semantics in Olfactory Perception},''
\newblock {\em ACL-IJCNLP 2015 - 53rd Annual Meeting of the Association for
  Computational Linguistics and the 7th International Joint Conference on
  Natural Language Processing of the Asian Federation of Natural Language
  Processing, Proceedings of the Conference, {\bf 2}}, pp.~231--236.

\bibitem{Blum1998CombiningCo-training}
Blum, A., and Mitchell, T., 1998,
\newblock ``{Combining labeled and unlabeled data with co-training},''
\newblock {\em Proceedings of the Annual ACM Conference on Computational
  Learning Theory}, pp.~92--100.

\bibitem{Levin2003UnsupervisedCo-training}
Levin, A., Viola, P., and Freund, Y., 2003,
\newblock ``{Unsupervised improvement of visual detectors using co-training},''
\newblock {\em Proceedings of the IEEE International Conference on Computer
  Vision, {\bf 1}}, pp.~626--633.

\bibitem{Christoudias2012Multi-ViewDisagreement}
Christoudias, C.~M., Urtasun, R., and Darrell, T., 2012,
\newblock ``{Multi-View Learning in the Presence of View Disagreement},''.

\bibitem{Girshick2015FastR-CNN}
Girshick, R., 2015,
\newblock ``{Fast R-CNN},''
\newblock In IEEE International Conference on Computer Vision (ICCV),
  pp.~1440--1448.

\bibitem{Cornia2022ExplainingAnalysis}
Cornia, M., Baraldi, L., and Cucchiara, R., 2022,
\newblock ``{Explaining transformer-based image captioning models: An empirical
  analysis},''
\newblock {\em AI Communications, {\bf 35}}(2), 1, pp.~111--129.

\bibitem{Herdade2019ImageWords}
Herdade, S., Kappeler, A., Boakye, K., and Soares, J., 2019,
\newblock ``{Image Captioning: Transforming Objects into Words},''
\newblock {\em Advances in Neural Information Processing Systems, {\bf 32}}.

\bibitem{Huang2019AttentionCaptioning}
Huang, L., Wang, W., Chen, J., and Wei, X.-Y., 2019,
\newblock ``{Attention on Attention for Image Captioning},''.

\bibitem{He2020ImageTransformer}
He, S., Liao, W., Tavakoli, H.~R., Yang, M., Rosenhahn, B., and Pugeault, N.,
  2020,
\newblock ``{Image Captioning through Image Transformer},''
\newblock {\em Lecture Notes in Computer Science (including subseries Lecture
  Notes in Artificial Intelligence and Lecture Notes in Bioinformatics), {\bf
  12625 LNCS}}, 4, pp.~153--169.

\bibitem{Li2019EntangledCaptioning}
Li, G., Zhu, L., Liu, P., and Yang, Y., 2019,
\newblock ``{Entangled transformer for image captioning},''
\newblock {\em Proceedings of the IEEE International Conference on Computer
  Vision, {\bf 2019-Octob}}, 10, pp.~8927--8936.

\bibitem{Aneja2017ConvolutionalCaptioning}
Aneja, J., Deshpande, A., and Schwing, A.~G., 2017,
\newblock ``{Convolutional Image Captioning},''
\newblock {\em Proceedings of the IEEE Computer Society Conference on Computer
  Vision and Pattern Recognition}, 11, pp.~5561--5570.

\bibitem{Deshpande2018FastPart-of-Speech}
Deshpande, A., Aneja, J., Wang, L., Schwing, A.~G., and Forsyth, D., 2018,
\newblock ``{Fast, Diverse and Accurate Image Captioning Guided By
  Part-of-Speech},''
\newblock {\em Proceedings of the IEEE Computer Society Conference on Computer
  Vision and Pattern Recognition, {\bf 2019-June}}, 5, pp.~10687--10696.

\bibitem{Li2019ControllableGeneration}
Li, B., Qi, X., Lukasiewicz, T., and Torr, P.~H., 2019,
\newblock ``{Controllable Text-to-Image Generation},''
\newblock {\em Advances in Neural Information Processing Systems, {\bf 32}}, 9.

\bibitem{Tao2020DF-GAN:Synthesis}
Tao, M., Tang, H., Wu, F., Jing, X.-Y., Bao, B.-K., and Xu, C., 2020,
\newblock ``{DF-GAN: A Simple and Effective Baseline for Text-to-Image
  Synthesis},''.

\bibitem{Karras2018ANetworks}
Karras, T., Laine, S., and Aila, T., 2018,
\newblock ``{A Style-Based Generator Architecture for Generative Adversarial
  Networks},''
\newblock {\em IEEE Transactions on Pattern Analysis and Machine Intelligence,
  {\bf 43}}(12), 12, pp.~4217--4228.

\bibitem{Patashnik2021StyleCLIP:Imagery}
Patashnik, O., Wu, Z., Shechtman, E., Cohen-Or, D., and Lischinski, D., 2021,
\newblock ``{StyleCLIP: Text-Driven Manipulation of StyleGAN Imagery},''
\newblock {\em Proceedings of the IEEE International Conference on Computer
  Vision}, 3, pp.~2065--2074.

\bibitem{Gal2021StyleGAN-NADA:Generators}
Gal, R., Patashnik, O., Maron, H., Bermano, A.~H., Chechik, G., and Cohen-Or,
  D., 2021,
\newblock ``{StyleGAN-NADA: CLIP-Guided Domain Adaptation of Image
  Generators},''
\newblock {\em ACM Transactions on Graphics, {\bf 41}}(4), 8.

\bibitem{Chefer2021Image-BasedTransfer}
Chefer, H., Benaim, S., Paiss, R., and Wolf, L., 2021,
\newblock ``{Image-Based CLIP-Guided Essence Transfer},''.

\bibitem{Ramesh2021Zero-ShotGeneration}
Ramesh, A., Pavlov, M., Goh, G., Gray, S., Voss, C., Radford, A., Chen, M., and
  Sutskever, I., 2021,
\newblock ``{Zero-Shot Text-to-Image Generation},''.

\bibitem{Crowson2022VQGAN-CLIP:Guidance}
Crowson, K., Biderman, S., Kornis, D., Stander, D., Hallahan, E., Castricato,
  L., Raff, E., and Allen~Hamilton, B., 2022,
\newblock ``{VQGAN-CLIP: Open Domain Image Generation and Editing with Natural
  Language Guidance},''.

\bibitem{Yu2021Vector-quantizedVQGAN}
Yu, J., Li, X., Koh, J.~Y., Zhang, H., Pang, R., Qin, J., Ku, A., Xu, Y.,
  Baldridge, J., and Wu, Y., 2021,
\newblock ``{Vector-quantized Image Modeling with Improved VQGAN},''.

\bibitem{Saharia2022PhotorealisticUnderstanding}
Saharia, C., Chan, W., Saxena, S., Li, L., Whang, J., Denton, E., Ghasemipour,
  S. K.~S., Ayan, B.~K., Mahdavi, S.~S., Lopes, R.~G., Salimans, T., Ho, J.,
  Fleet, D.~J., and Norouzi, M., 2022,
\newblock ``{Photorealistic Text-to-Image Diffusion Models with Deep Language
  Understanding},''.

\bibitem{Frans2021CLIPDraw:Encoders}
Frans, K., Soros, L.~B., and Witkowski, O., 2021,
\newblock ``{CLIPDraw: Exploring Text-to-Drawing Synthesis through
  Language-Image Encoders},''.

\bibitem{Choy20163D-R2N2:Reconstruction}
Choy, C.~B., Xu, D., Gwak, J.~Y., Chen, K., and Savarese, S., 2016,
\newblock ``{3D-R2N2: A unified approach for single and multi-view 3D object
  reconstruction},''
\newblock {\em Lecture Notes in Computer Science (including subseries Lecture
  Notes in Artificial Intelligence and Lecture Notes in Bioinformatics), {\bf
  9912 LNCS}}, pp.~628--644.

\bibitem{Gkioxari2019MeshR-CNN}
Gkioxari, G., Johnson, J., and Malik, J., 2019,
\newblock ``{Mesh R-CNN},''
\newblock {\em Proceedings of the IEEE International Conference on Computer
  Vision, {\bf 2019-Octob}}, 6, pp.~9784--9794.

\bibitem{Shrestha2020MeshMVS:Reconstruction}
Shrestha, R., Fan, Z., Su, Q., Dai, Z., Zhu, S., and Tan, P., 2020,
\newblock ``{MeshMVS: Multi-View Stereo Guided Mesh Reconstruction},''
\newblock {\em Proceedings - 2021 International Conference on 3D Vision, 3DV
  2021}, 10, pp.~1290--1300.

\bibitem{Fan2016AImage}
Fan, H., Su, H., and Guibas, L., 2016,
\newblock ``{A Point Set Generation Network for 3D Object Reconstruction from a
  Single Image},''
\newblock {\em Proceedings - 30th IEEE Conference on Computer Vision and
  Pattern Recognition, CVPR 2017, {\bf 2017-Janua}}, 12, pp.~2463--2471.

\bibitem{Groueix2018AtlasNet:Generation}
Groueix, T., Fisher, M., Kim, V.~G., Russell, B.~C., and Aubry, M., 2018,
\newblock ``{AtlasNet: A Papier-M{\textbackslash}{\^{}}ach{\textbackslash}'e
  Approach to Learning 3D Surface Generation},''.

\bibitem{Li2022AAutoencoder}
Li, X., Xie, C., and Sha, Z., 2022,
\newblock ``{A Predictive and Generative Design Approach for Three-Dimensional
  Mesh Shapes Using Target-Embedding Variational Autoencoder},''
\newblock {\em Journal of Mechanical Design, {\bf 144}}(11), 11.

\bibitem{Wu2016LearningModeling}
Wu, J., Zhang, C., Xue, T., Freeman, W.~T., and Tenenbaum, J.~B., 2016,
\newblock ``{Learning a Probabilistic Latent Space of Object Shapes via 3D
  Generative-Adversarial Modeling},''
\newblock {\em Advances in Neural Information Processing Systems}, 10,
  pp.~82--90.

\bibitem{Khan2019UnsupervisedModeling}
Khan, S.~H., Guo, Y., Hayat, M., and Barnes, N., 2019,
\newblock ``{Unsupervised Primitive Discovery for Improved 3D Generative
  Modeling},''
\newblock {\em Proceedings of the IEEE Computer Society Conference on Computer
  Vision and Pattern Recognition, {\bf 2019-June}}, 6, pp.~9731--9740.

\bibitem{Maron2017ConvolutionalCovers}
Maron, H., Galun, M., Aigerman, N., Trope, M., Dym, N., Yumer, E., Kim, V.~G.,
  and Lipman, Y., 2017,
\newblock ``{Convolutional neural networks on surfaces via seamless toric
  covers},''
\newblock {\em ACM Transactions on Graphics (TOG), {\bf 36}}(4), 7.

\bibitem{Ben-Hamu2018Multi-chartModeling}
Ben-Hamu, H., Maron, H., Kezurer, I., Avineri, G., and Lipman, Y., 2018,
\newblock ``{Multi-chart Generative Surface Modeling},''
\newblock {\em SIGGRAPH Asia 2018 Technical Papers, SIGGRAPH Asia 2018}, 6.

\bibitem{Saquil2020Rank3DGAN:Attributes}
Saquil, Y., Xu, Q.~C., Yang, Y.~L., and Hall, P., 2020,
\newblock ``{Rank3DGAN: Semantic mesh generation using relative attributes},''
\newblock {\em AAAI 2020 - 34th AAAI Conference on Artificial Intelligence},
  pp.~5586--5594.

\bibitem{Alhaija2022XDGAN:Space}
Alhaija, H.~A., Dirik, A., Kn{\"{o}}rig, A., Fidler, S., and Shugrina, M.,
  2022,
\newblock ``{XDGAN: Multi-Modal 3D Shape Generation in 2D Space},''.

\bibitem{Fu2022ShapeCrafter:Model}
Fu, R., Zhan, X., Chen, Y., Ritchie, D., and Sridhar, S., 2022,
\newblock ``{ShapeCrafter: A Recursive Text-Conditioned 3D Shape Generation
  Model},''.

\bibitem{Kalischek2022TetrahedralGeneration}
Kalischek, N., Peters, T., Wegner, J.~D., and Schindler, K., 2022,
\newblock ``{Tetrahedral Diffusion Models for 3D Shape Generation},''.

\bibitem{Alwala2022Pre-trainReconstruction}
Alwala, K.~V., Gupta, A., and Tulsiani, S., 2022,
\newblock ``{Pre-train, Self-train, Distill: A simple recipe for Supersizing 3D
  Reconstruction},''
\newblock pp.~3763--3772.

\bibitem{Liu2022ISS:Generation}
Liu, Z., Dai, P., Li, R., Qi, X., and Fu, C.-W., 2022,
\newblock ``{ISS: Image as Stepping Stone for Text-Guided 3D Shape
  Generation},''.

\bibitem{Nam20223D-LDM:Models}
Nam, G., Khlifi, M., Rodriguez, A., Tono, A., Zhou, L., and Guerrero, P., 2022,
\newblock ``{3D-LDM: Neural Implicit 3D Shape Generation with Latent Diffusion
  Models},''.

\bibitem{Cheng2022Cross-ModalManipulation}
Cheng, Z., Chai, M., Ren, J., Lee, H.-Y., Olszewski, K., Huang, Z., Maji, S.,
  and Tulyakov, S., 2022,
\newblock ``{Cross-Modal 3D Shape Generation and Manipulation},''
\newblock pp.~303--321.

\bibitem{Wang2018Pixel2Mesh:Images}
Wang, N., Zhang, Y., Li, Z., Fu, Y., Liu, W., and Jiang, Y.~G., 2018,
\newblock ``{Pixel2Mesh: Generating 3D Mesh Models from Single RGB Images},''
\newblock {\em Lecture Notes in Computer Science (including subseries Lecture
  Notes in Artificial Intelligence and Lecture Notes in Bioinformatics), {\bf
  11215 LNCS}}, 4, pp.~55--71.

\bibitem{Michel2021Text2Mesh:Meshes}
Michel, O., Bar-On, R., Liu, R., Benaim, S., and Hanocka, R., 2021,
\newblock ``{Text2Mesh: Text-Driven Neural Stylization for Meshes},''.

\bibitem{Jetchev2021ClipMatrix:Meshes}
Jetchev, N., 2021,
\newblock ``{ClipMatrix: Text-controlled Creation of 3D Textured Meshes},''.

\bibitem{hycon}
Mai, S., Zeng, Y., Zheng, S., and Hu, H., 2022,
\newblock ``Hybrid contrastive learning of tri-modal representation for
  multimodal sentiment analysis,''
\newblock {\em IEEE Transactions on Affective Computing}, pp.~1--1.

\bibitem{fnd-clip}
Zhou, Y., Ying, Q., Qian, Z., Li, S., and Zhang, X., 2022,
\newblock Multimodal fake news detection via clip-guided learning.

\bibitem{ddimdl}
Deng, Y., Xu, X., Qiu, Y., Xia, J., Zhang, W., and Liu, S., 2020,
\newblock ``{A multimodal deep learning framework for predicting drug–drug
  interaction events},''
\newblock {\em Bioinformatics, {\bf 36}}(15), 05, pp.~4316--4322.

\bibitem{deeptake}
Pakdamanian, E., Sheng, S., Baee, S., Heo, S., Kraus, S., and Feng, L., 2021,
\newblock ``Deeptake: Prediction of driver takeover behavior using multimodal
  data,''
\newblock In Proceedings of the 2021 CHI Conference on Human Factors in
  Computing Systems, CHI '21, Association for Computing Machinery.

\bibitem{10.1115/1.4056500}
Yuan, C., Marion, T., and Moghaddam, M., 2023,
\newblock ``{DDE-GAN: Integrating a Data-Driven Design Evaluator Into
  Generative Adversarial Networks for Desirable and Diverse Concept
  Generation},''
\newblock {\em Journal of Mechanical Design, {\bf 145}}(4), p.~041407.

\bibitem{Ordonez2011Im2Text:Photographs}
Ordonez, V., Kulkarni, G., and Berg, T.~L., 2011,
\newblock ``{Im2Text: Describing Images Using 1 Million Captioned
  Photographs},''
\newblock In NIPS 12.

\bibitem{Devlin2015LanguageWorks}
Devlin, J., Cheng, H., Fang, H., Gupta, S., Deng, L., He, X., Zweig, G., and
  Mitchell, M., 2015,
\newblock ``{Language Models for Image Captioning: The Quirks and What
  Works},''
\newblock {\em ACL-IJCNLP 2015 - 53rd Annual Meeting of the Association for
  Computational Linguistics and the 7th International Joint Conference on
  Natural Language Processing of the Asian Federation of Natural Language
  Processing, Proceedings of the Conference, {\bf 2}}, 5, pp.~100--105.

\bibitem{Kwon2022EnablingLearning}
Kwon, E., Huang, F., and Goucher-Lambert, K., 2022,
\newblock ``{Enabling multi-modal search for inspirational design stimuli using
  deep learning},''
\newblock {\em AI EDAM, {\bf 36}}, 7, p.~e22.

\bibitem{Farhadi2010EveryImages}
Farhadi, A., Hejrati, M., Sadeghi, M.~A., Young, P., Rashtchian, C.,
  Hockenmaier, J., and Forsyth, D., 2010,
\newblock ``{Every picture tells a story: Generating sentences from images},''
\newblock {\em Lecture Notes in Computer Science (including subseries Lecture
  Notes in Artificial Intelligence and Lecture Notes in Bioinformatics), {\bf
  6314 LNCS}}(PART 4), pp.~15--29.

\bibitem{Xu2015JointlyFramework}
Xu, R., Xiong, C., Chen, W., and Corso, J.~J., 2015,
\newblock ``{Jointly Modeling Deep Video and Compositional Text to Bridge
  Vision and Language in a Unified Framework},''
\newblock {\em Proceedings of the AAAI Conference on Artificial Intelligence,
  {\bf 29}}(1), 2, pp.~2346--2352.

\bibitem{Hodosh2013FramingMetrics}
Hodosh, M., Young, P., and Hockenmaier, J., 2013,
\newblock ``{Framing Image Description as a Ranking Task: Data, Models and
  Evaluation Metrics},''
\newblock {\em Journal of Artificial Intelligence Research, {\bf 47}}, 8,
  pp.~853--899.

\bibitem{Gero2019TheModel}
Gero, J., and Milovanovic, J., 2019,
\newblock ``{The situated function-behavior-structure co-design model},''
\newblock {\em https://doi.org/10.1080/15710882.2019.1654524, {\bf 17}}(2),
  pp.~211--236.

\bibitem{Lin2014MicrosoftContext}
Lin, T.~Y., Maire, M., Belongie, S., Hays, J., Perona, P., Ramanan, D.,
  Doll{\'{a}}r, P., and Zitnick, C.~L., 2014,
\newblock ``{Microsoft COCO: Common Objects in Context},''
\newblock {\em Lecture Notes in Computer Science (including subseries Lecture
  Notes in Artificial Intelligence and Lecture Notes in Bioinformatics), {\bf
  8693 LNCS}}(PART 5), 5, pp.~740--755.

\bibitem{Krishna2016VisualAnnotations}
Krishna, R., Zhu, Y., Groth, O., Johnson, J., Hata, K., Kravitz, J., Chen, S.,
  Kalantidis, Y., Li, L.~J., Shamma, D.~A., Bernstein, M.~S., and Fei-Fei, L.,
  2016,
\newblock ``{Visual Genome: Connecting Language and Vision Using Crowdsourced
  Dense Image Annotations},''
\newblock {\em International Journal of Computer Vision, {\bf 123}}(1), 2,
  pp.~32--73.

\bibitem{Thomee2015YFCC100M:Research}
Thomee, B., Shamma, D.~A., Friedland, G., Elizalde, B., Ni, K., Poland, D.,
  Borth, D., and Li, L.-J., 2015,
\newblock ``{YFCC100M: The New Data in Multimedia Research},''
\newblock {\em Communications of the ACM, {\bf 59}}(2), 3, pp.~64--73.

\bibitem{Sun2017RevisitingEra}
Sun, C., Shrivastava, A., Singh, S., and Gupta, A., 2017,
\newblock ``{Revisiting Unreasonable Effectiveness of Data in Deep Learning
  Era},''
\newblock {\em Proceedings of the IEEE International Conference on Computer
  Vision, {\bf 2017-Octob}}, 7, pp.~843--852.

\bibitem{Murray2012AVA:Analysis}
Murray, N., Marchesotti, L., and Perronnin, F., 2012,
\newblock ``{AVA: A large-scale database for aesthetic visual analysis},''
\newblock {\em Proceedings of the IEEE Computer Society Conference on Computer
  Vision and Pattern Recognition}, pp.~2408--2415.

\bibitem{Chen2018Text2Shape:Embeddings}
Chen, K., Choy, C.~B., Savva, M., Chang, A.~X., Funkhouser, T., and Savarese,
  S., 2018,
\newblock ``{Text2Shape: Generating Shapes from Natural Language by Learning
  Joint Embeddings},''
\newblock {\em Lecture Notes in Computer Science (including subseries Lecture
  Notes in Artificial Intelligence and Lecture Notes in Bioinformatics), {\bf
  11363 LNCS}}, 12, pp.~100--116.

\bibitem{Jahan2021ParkinsonsLearning}
Jahan, N., Nesa, A., and Layek, M.~A., 2021,
\newblock ``{Parkinson's Disease Detection Using CNN Architectures with
  Transfer Learning},''
\newblock In 2021 International Conference on Innovative Computing, Intelligent
  Communication and Smart Electrical Systems (ICSES), Institute of Electrical
  and Electronics Engineers (IEEE), pp.~1--5.

\bibitem{Regenwetter2023BeyondDesign}
Regenwetter, L., Srivastava, A., Gutfreund, D., and Ahmed, F., 2023,
\newblock ``{Beyond Statistical Similarity: Rethinking Metrics for Deep
  Generative Models in Engineering Design},''.

\bibitem{Xu2022DeepSurvey}
Xu, P., Hospedales, T.~M., Yin, Q., Song, Y.-Z., Xiang, T., and Wang, L., 2022,
\newblock ``{Deep Learning for Free-Hand Sketch: A Survey},''
\newblock {\em IEEE transactions on pattern analysis and machine intelligence},
  1.

\bibitem{Ghadai2019Multi-levelFeatures}
Ghadai, S., Lee, X.~Y., Balu, A., Sarkar, S., and Krishnamurthy, A., 2019,
\newblock ``{Multi-level 3D CNN for learning multi-scale spatial features},''
\newblock {\em IEEE Computer Society Conference on Computer Vision and Pattern
  Recognition Workshops, {\bf 2019-June}}, 6, pp.~1152--1156.

\bibitem{Kong2014WhatCoreference}
Kong, C., Lin, D., Bansal, M., Urtasun, R., and Fidler, S., 2014,
\newblock ``{What are you talking about? Text-to-image coreference},''
\newblock {\em Proceedings of the IEEE Computer Society Conference on Computer
  Vision and Pattern Recognition}, 9, pp.~3558--3565.

\bibitem{Wu2022ResearchCOVID-19}
Wu, F., Lin, Y.~C., and Lu, P., 2022,
\newblock ``{Research on the Design Strategy of Healing Products for Anxious
  Users during COVID-19},''
\newblock {\em International Journal of Environmental Research and Public
  Health, {\bf 19}}(10), 5.

\bibitem{Linardatos2021ExplainableMethods}
Linardatos, P., Papastefanopoulos, V., and Kotsiantis, S., 2021,
\newblock ``{Explainable AI: A Review of Machine Learning Interpretability
  Methods},''
\newblock {\em Entropy, {\bf 23}}(1), 1, pp.~1--45.

\bibitem{BarredoArrieta2020ExplainableAI}
Barredo~Arrieta, A., D{\'{i}}az-Rodr{\'{i}}guez, N., Del~Ser, J., Bennetot, A.,
  Tabik, S., Barbado, A., Garcia, S., Gil-Lopez, S., Molina, D., Benjamins, R.,
  Chatila, R., and Herrera, F., 2020,
\newblock ``{Explainable Explainable Artificial Intelligence (XAI): Concepts,
  taxonomies, opportunities and challenges toward responsible AI},''
\newblock {\em Information Fusion, {\bf 58}}, 6, pp.~82--115.

\end{thebibliography}



\end{document}